%% file: main.tex
\documentclass[nohyperref]{article}

\PassOptionsToPackage{numbers, compress}{natbib}
\usepackage[preprint]{neurips}

\usepackage[T1]{fontenc}    
\usepackage{hyperref}       
\usepackage{url}            
\usepackage{graphicx}    
\usepackage{booktabs}  
\usepackage{amsmath}
\usepackage{amsfonts} 
\usepackage{dsfont}
\usepackage{wrapfig}
\usepackage{graphicx}
\usepackage{amsthm}
\usepackage{amssymb}
\usepackage{nicefrac}       
\usepackage{microtype}      
\usepackage{lipsum}
\usepackage{graphicx}
\usepackage{caption}
\input{math_symbols.tex}

\usepackage{xcolor}
\usepackage{amsmath}
\newtheorem{defn}{Definition}

\newtheorem{theorem}[defn]{Theorem}

\newtheorem{proposition}[defn]{Proposition}

\usepackage[american]{babel}

\usepackage{natbib} 
\usepackage{mathtools}
\newcommand{\shortparagraph}[1]{\textbf{#1}\quad}

\global\long\def\rkhs{\mathcal{H}_k}
\global\long\def\calX{\mathcal{X}}
\global\long\def\calY{\mathcal{Y}}

\usepackage[most]{tcolorbox}

\definecolor{olive}{rgb}{0.6, 0.6, 0.2}
\definecolor{sand}{rgb}{0.8666666666666667, 0.8, 0.4666666666666667}
\definecolor{wine}{rgb}{0.5333333333333333, 0.13333333333333333, 0.3333333333333333}
\definecolor{deblue}{RGB}{11,132,147}
\definecolor{ocra}{RGB}{204, 119, 34}

\newtcolorbox{CatchyBox}[2][]{
    lower separated=false,
    colback=white!90!blue!90!ocra,
    colframe=white, fonttitle=\bfseries,
    colbacktitle=white!70!blue!90!ocra,
    coltitle=black,
    enhanced,
    attach boxed title to top left={xshift=.02\linewidth,yshift=-4mm},
    title=#2,#1}

\usepackage{subcaption}
\begin{document}

\title{\textsc{RKHS-SHAP}: Shapley Values for Kernel Methods}

\author{%
Siu Lun Chau \\
Department of Statistics \\
University of Oxford \\
\And
Robert Hu \\
Department of Statistics \\
University of Oxford \\
\And
Javier Gonzalez \\
Microsoft Research Cambridge\\
Cambridge \\
\And 
Dino Sejdinovic \\
Department of Statistics \\
University of Oxford \\

}

\maketitle

\begin{abstract}
    Feature attribution for kernel methods is often heuristic and not individualised for each prediction. To address this, we turn to the concept of Shapley values~(SV), a coalition game theoretical framework that has previously been applied to different machine learning model interpretation tasks, such as linear models, tree ensembles and deep networks. By analysing SVs from a functional perspective, we propose \textsc{RKHS-SHAP}, an attribution method for kernel machines that can efficiently compute both \emph{Interventional} and \emph{Observational Shapley values} using kernel mean embeddings of distributions. We show theoretically that our method is robust with respect to local perturbations - a key yet often overlooked desideratum for consistent model interpretation. Further, we propose \emph{Shapley regulariser}, applicable to a general empirical risk minimisation framework, allowing learning while controlling the level of specific feature's contributions to the model. We demonstrate that the Shapley regulariser enables learning which is robust to covariate shift of a given feature and fair learning which controls the SVs of sensitive features. 
\end{abstract}

\input{sections/01-introduction}

\input{sections/02-background-materials}
\input{sections/03-main-method}

\input{sections/04-regulariser}

\input{sections/05-experiment-new}

\input{sections/06-discussion}

\clearpage
\bibliography{main}

\input{sections/07-appendix}
\end{document}

%% file: math_symbols.tex
\def\RR{{\mathbb R}}    
\def\EE{{\mathbb E}}    
\def\11{{\mathbf 1}}    

     \def\cB{{\mathcal B}}  \def\cH{{\mathcal H}}      \def\cO{{\mathcal O}}         \def\cK{{\mathcal K}}     \def\cL{{\mathcal L}}  \def\cX{{\mathcal X}} \def\cY{{\mathcal Y}}  

\def\bfx{{\bf x}} \def\bfy{{\bf y}} \def\bfz{{\bf z}} 
 \def\bfK{{\bf K}}

\def\balpha{\boldsymbol{\alpha}}



%% file: sections/01-introduction.tex
\vspace{-0.6cm}
\section{Introduction}
\vspace{-0.2cm}

Machine learning model interpretability is critical for researchers, data scientists, and developers to explain, debug and trust their models and understand the value of their findings. A typical way to understand model performance is to attribute importance scores to each input feature~\citep{carvalho2019machine}. These scores can be computed either for an entire dataset to explain the model's overall behaviour (global) or compute individually for each single prediction (local).

\begin{wrapfigure}[13]{r}{0.35\textwidth}
  \begin{center}
  \vspace{-1.2cm}
    \includegraphics[width=0.35\textwidth]{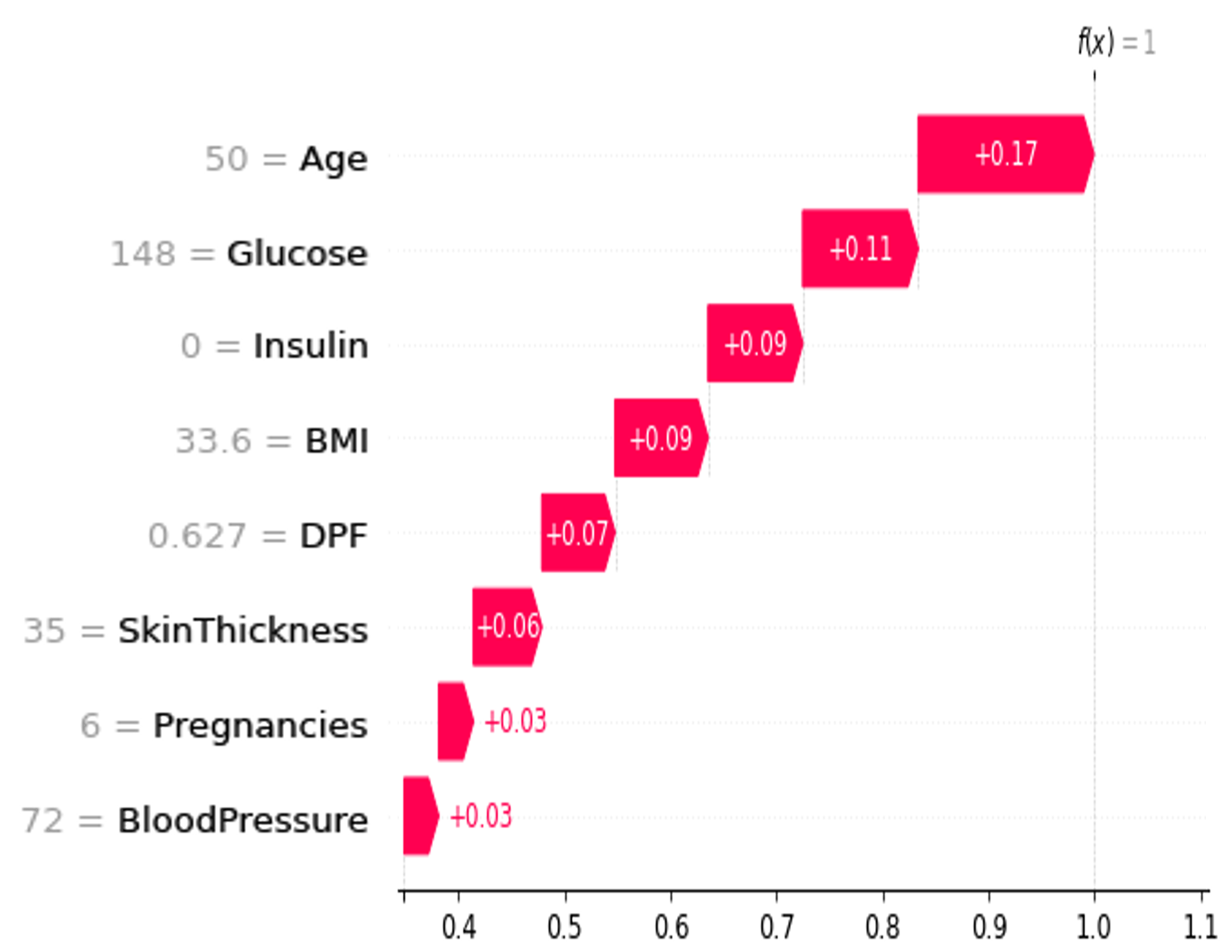}
  \end{center}
  \vspace{-0.2cm}
    \caption{\small{An example of \textsc{RKHS-SHAP} providing local explanations to why a kernel logistic model predicts this patient to be diabetic~\cite{diabetes_data}.}}
  \label{fig: illustration}
\end{wrapfigure}

Understanding feature importances in reproducing kernel Hilbert space (RKHS) methods such as kernel ridge regression and support vector machines often require the study of kernel lengthscales across dimensions~\citep[Chapter 5]{williams2006gaussian}. The larger the value, the less relevant the feature is to the model. Albeit straightforward, this approach comes with three shortcomings: (1) It only provides global feature importances and cannot be individualised to each single prediction. This explanation is limited as global importance does not necessarily imply local importance~\cite{lime}). In safety critical domain such as medicine, understanding individual prediction is arguably more important than capturing the general model performance. See Fig~\ref{fig: illustration} for an example of local explanation. (2) The tuning of lengthscales often requires a user-specified grid of possible configurations and is selected using cross-validations. This pre-specification thus injects substantial amount of human bias to the explanation task.  (3) Lengthscales across kernels acting on different data types, such as binary and continuous variables, are difficult to compare and interpret.

To address this problem we turn to the Shapley value (SV)~\citep{shapley1953value} literature, which has become central to many model explanation methods in recent years. The Shapley value was originally a concept used in game theory that involves fairly distributing credits to players working in coalition.~\citet{vstrumbelj2014explaining} were one of the first to connect SV with machine learning explanations by casting predictions as coalition games, and features as players. Since then, a variety of SV based explanation models were proposed. For example, \textsc{LinearSHAP}~\citep{vstrumbelj2014explaining} for linear models, \textsc{TreeSHAP}~\citep{lundberg2018consistent} for tree ensembles and \textsc{DeepSHAP}~\citep{lundberg2017unified} for deep networks. Model agnostic methods such as \textsc{Data-Shapley}~\citep{ghorbani2019data}, \textsc{SAGE}~\citep{covert2020understanding} and \textsc{KernelSHAP} \footnote{The kernel in \textsc{KernelSHAP} refers to the estimation procedure is not related to RKHS kernel methods.}~\citep{lundberg2017unified} were also proposed. However, to the best of our knowledge, an SV-based local feature attribution framework suited for kernel methods has not been proposed. 

While one could still apply model-agnostic \textsc{KernelSHAP} on kernel machines, we show that by  representing distributions as elements in the RKHS through kernel mean embeddings~\citep{song2013kernel,muandet2016kernel}, we can compute Shapley values more efficiently by circumventing the need to sample and estimate an exponential amount of densities. We call this approach \textsc{RKHS-SHAP} to distinguish it from \textsc{KernelSHAP}. Through the lens of RKHS, we study Shapley values from a functional perspective and prove that our method is robust with respect to local perturbations under mild assumptions, which is an important yet often neglected criteria for explanation models as discussed in~\citet{hancox2020robustness}. In addition, a \emph{Shapley regulariser} based on \textsc{RKHS-SHAP} is proposed for the empirical risk minimisation framework, allowing the modeller to control the degree of feature contribution during the learning. We also discuss its application to robust learning to covariate shift of a given feature and fair learning while controlling contributions from sensitive features. We summarise our contributions below:

\textbf{1.} \quad We propose \textsc{RKHS-SHAP}, a model specific algorithm to compute Shapley values efficiently for kernel methods by circumventing the need to sample and fit from an exponential number of densities.

\vspace{-0.1cm}

\textbf{2.} \quad We prove that the corresponding Shapley values are robust to local perturbations under mild assumptions, thus providing consistent explanations for the kernel model.

\vspace{-0.1cm}

\textbf{3.} \quad We propose a \emph{Shapley regulariser} for the empirical risk minimisation framework, allowing the modeller to control the degree of feature contribution during the learning.


The paper is outlined as follows: In section 2 we provide an overview of Shapley values and kernel methods. In section 3 we introduce \textsc{RKHS-SHAP} and show robustness of the algorithm. \emph{Shapley regulariser} is introduced in section 4. Section 5 provides extensive experiments. We conclude our work in section 6.

%% file: sections/02-background-materials.tex
\vspace{-0.3cm}
\section{Background Materials}
\vspace{-0.2cm}

\textbf{Notation.} \quad We denote $X, Y$ as random variables (rv) with distribution $p(X, Y)$ taking values in the $d$-dimensional instance space $\cX \subseteq \RR^d$ and the label space $\cY$ (could be in $\RR$ or discrete) respectively. We use $D = \{1,...,d\}$ to denote the feature index set of $X$ and $S \subseteq D$ to denote the subset of features of interests. Lower case letters are used to denote observations from corresponding rvs.

\vspace{-0.2cm}
\subsection{The Shapley Value}
\vspace{-0.1cm}

The Shapley value was first proposed by~\citet{shapley1953value} to allocate performance credit across coalition game players in the following sense: Let $\nu: \{0, 1\}^d \to \RR$ be a \emph{coalition game} that returns a score for each coalition $S \subseteq D_g$, where $D_g=\{1, ..., d\}$ represents a set of players. Assuming the grand coalition $D_g$ is participating and one wished to provide the $i^{th}$ player with a fair allocation of the total profit $\nu(D_g)$, how should one do it? Surely this is related to each player's \emph{marginal contribution} to the profit with respect to a coalition $S$, i.e. $\nu(S\cup i) - \nu(S)$. \citet{shapley1953value} proved that there exists a \emph{unique} combination of marginal contributions that satisfies a set of favourable and fair game theoretical axioms, commonly known as \emph{efficiency, null player, symmetry} and \emph{additivity}. This unique combination of contributions is later denoted as the \emph{Shapley value}. Formally, given a coalition game $\nu$, the Shapley value for player $i$ is computed as the following,
\begin{align}
\small
\label{eq: shapley value}
\phi_i(\nu) = \frac{1}{d} \sum_{S\subseteq D_g\backslash \{i\}} {d-1 \choose |S|}^{-1} \Big(\nu(S\cup i) - \nu(S) \Big).
\end{align}




\textbf{Choosing $\nu$ for ML explanation}\quad In recent years, the Shapley value concept has become popular for feature attribution in machine learning. \textsc{SHAP}~\citep{lundberg2017unified}, \textsc{Shapley effect}~\citep{song2016shapley}, \textsc{Data-Shapley}~\citep{ghorbani2019data} and \textsc{SAGE}~\citep{covert2020understanding} are all examples that cast model explanations as coalition games by choosing problem-specific value functions $\nu$. Denote $f: \cX \to \cY$ as the machine learning model of interest. Value functions for local attribution on observation $x$ often take the form of the expectation of $f$ with respect to some reference distribution $r(X_{S^c}\mid X_S = x_S)$, where $S\subseteq D$ is some coalition of features in analogous to the game theory setting, such that:
\begin{align}
    \nu_{x, S}(f) = \EE_{r(X_{S^c}\mid X_S=x_S)}[f(\{x_S, X_{S^c}\}],
\end{align}
where $\{x_S, X_{S^c}\}$ denotes the concatenation of the arguments. We wrote $f$ as the main argument of $\nu$ to highlight its interpretation as a functional indexed by local observation $x$ and coalition $S$. When $r$ is set to be marginal distribution, i.e $r(X_{S^c}\mid X_S = x_S) = p(X_{S^c})$, the value function is denoted as the \emph{Interventional value function} by~\citet{janzing2020feature}. \emph{Observational value function}~\citep{frye2020shapley}, on the other hand, set the reference distribution to be a conditional distribution $p(X_{S^c}\mid X_S=x_S)$.  Other choices of reference distributions will lead to Shapley values with specific properties, e.g., better locality of explanations~\citep{ghalebikesabi2021locality} or incorporating causal knowledge~\citep{heskes2020causal}. In this work we shall restrict our attention to marginal and conditional cases as they are the two most commonly adopted choices in the literature.


\begin{defn} Given model $f$, local observation $x$ and a coalition set $S\subseteq D$, the Interventional and Observational value functions are denoted by $\nu_{x, S}^{(I)}(f) := \EE_{X_{S^c}}[(f(\{x_S, X_{S^c}\})]$ and $\nu_{x, S}^{(O)}(f) := \EE_{X_{S^c}}[f(\{x_S, X_{S^c}\})\mid X_S=x_S]$. 
\end{defn}

The right choice of $\nu$ has been a long-standing debate in the community. While~\citet{janzing2020feature} argued from a causal perspective that $\nu_{x, f}^{(I)}$ is the correct notion to represent missingness of features in an explanation task,~\citet{frye2020shapley} argued that computing marginal expectation ignores feature correlation and leads to unrealistic results since one would be evaluating the value function outside the data-manifold. This controversy was further investigated by~\citet{chen2020true}, where they argued that the choice of $\nu$ is \emph{application dependent} and the two approaches each lead to an explanation that is either \emph{true to the model} (marginal expectation) or \emph{true to the data} (conditional expectation). When the context is clear, we denote the Shapley value of the $i^{th}$ feature of observation $x$ at $f$ as $\phi_{x, i}(f)$ and use a superscript to indicate whether it is \emph{Interventional} $\phi_{x, i}^{(I)}(f)$ or \emph{Observational} $\phi_{x, i}^{(O)}(f)$.


\shortparagraph{Computing Shapley values.} While Shapley values can be estimated directly from Eq. \eqref{eq: shapley value} using a sampling approach~\citep{vstrumbelj2014explaining},~\citet{lundberg2017unified} proposed \textsc{KernelSHAP}, a more efficient algorithm for estimating Shapley values in high dimensional feature spaces by casting Eq. \eqref{eq: shapley value} as a weighted least square problem. Similar to \textsc{LIME}~\citep{lime}, for each data $x$, model $f$, and feature coalition $S$, \textsc{KernelSHAP} places a linear model $u_x(S) = \beta_{x,0} + \sum_{i\in S} \beta_{x,i}$ to explain the value function $\nu_{x, S}(f)$, which corresponds to solving the following regression problem: $\min_{\beta_{x,0}, ..., \beta_{x,d}}\sum_{S\subseteq D} w(S) (u_x(S) - \nu_{x, S}(f) )^2$, where $w(S) = \frac{d-1}{{d \choose |S|}|S|(d-|S|)}$ is a carefully chosen weighting such that the regression coefficients recover Shapley values. In particular, one set $w(\varnothing) = w(D) = \infty$ to effectively enforce constraints $\beta_{x,0} = \nu_{x, \varnothing}(f)$ and $\sum_{i\in D}\beta_{x,i} = \nu_{x,D}(f)-\nu_{x, \varnothing}(f)$. Denoting each subset $S\subseteq D$ using the corresponding binary vector $\bfz\in \{0, 1\}^d$, and with an abuse of notation by setting $\nu_{\cdot, \bfz} := \nu_{\cdot, S}$ and $w(\bfz) := w(S)$ for $S = \{j: \bfz[j] = 1\}$, we can express the Shapley values $\boldsymbol{\beta}_x := [\beta_{x,0}, ..., \beta_{x,d}]$ as $\boldsymbol{\beta}_x = (Z^\top W Z)^{-1}Z^\top W \mathbf{v}_x$ where $Z \in \RR^{2^d \times d}$ is the binary matrices with columns $\{\bfz_i\}_{i=1}^{2^d}$, $W$ is the diagonal matrix with entries $w_{ii} = w(\bfz_i)$ and ${\bf v}_x := \{\nu_{x, \bfz_i}(f)\}_{i=1}^{2^d} \in \RR^{2^d\times 1}$ the vector of evaluated value functions, which is often estimated using sampling and data imputations. We shall explain the pathology of this approach in detail later in Section~\ref{sec: method}. In practice, instead of evaluating at all $2^d$ combinations, one would subsample the coalitions $z \sim w(z)$ for computational efficiency~\citep{covert2021improving}. 

\label{subsec: model-specific-sv}
\shortparagraph{Model specific Shapley methods.} \textsc{KernelSHAP} provides efficient model-agnostic estimations of Shapley values. However, by leveraging additional structural knowledge about specific models, one could further improve computational performance. This leads to a variety of model-specific approximations, most of which relies on utilising their specific structure to speed up computation of value functions. For example, \textsc{LinearSHAP}~\citep{vstrumbelj2014explaining} explain linear models using model coefficients directly. \textsc{TreeSHAP}~\citep{lundberg2018consistent} provides an exponential reduction in complexity compared to \textsc{KernelSHAP} by exploiting the tree structure. \textsc{DeepSHAP}~\citep{lundberg2017unified}, on the other hand, combines \textsc{DeepLIFT}~\citep{shrikumar2017learning} with Shapley values and uses the compositional nature of deep networks to improve efficiencies. However, to the best of our knowledge, a kernel method specific Shapley value approximation has not been studied. Later in Section~\ref{sec: method}, we will show that, under a mild structural assumption on the RKHS, kernel methods can be used to speed up the computation in \textsc{KernelSHAP} by estimating value functions analytically, thus circumventing the need for estimating and sampling from an exponential number of densities. 


\textbf{Related work on kernel-based Shapley methods.}\quad \citet{da2021kernel}'s work on tackling global sensitive analysis by proposing the kernel-based maximum mean discrepancy as value function, is conceptually most similar to ours. However, there are multiple key differences in our contributions. Firstly, their method is designed for global explanation, while ours is for local. Secondly, similar to interventional SV, they do not consider any conditional distributions, thus leading to completely different estimation procedures and thus novelty. Lastly, their method is on understanding the input/outputs relationship of a numerical simulation model, while ours focuses on understanding specific RKHS models learnt from a machine learning task, e.g. kernel ridge regression and kernel logistic regression.

\vspace{-0.3cm}
\subsection{Kernel Methods}
\label{sec: kernel methods}
\vspace{-0.2cm}
Kernel methods are one of the pillars of machine learning, as they provide flexible yet principled ways to model complex functional relationships and come with well-established statistical properties and theoretical guarantees. 


\textbf{Empirical Risk Minimisation.}\quad Recall in the supervised learning framework, we are learning a function $f:\cX \rightarrow \cY$ from a hypothesis space $\cH$, such that given a training set $(\bfx, \bfy) = \{(x_i, y_i)\}_{i=1}^n$ sampled identically and independently from $p$, the following empirical risk is minimised: $f^* = \arg\min_{f \in \cH} \frac{1}{n}\sum_{i=1}^n \ell(y_i, f(x_i)) + \lambda_f \Omega(f)$,
where $\ell:\cY \times \cY \rightarrow \RR$ is the loss function, $\Omega: \cH \rightarrow \RR$ a regularisation function and $\lambda_f$ a scalar controlling the level of regularisation. Denote $k:\cX\times \cX\rightarrow\RR$ a positive definite kernel with feature map $\psi_x$ for input $x\in \calX$ and $\cH_k$ the corresponding RKHS. If we pick $\cH_k$ as our hypothesis space, then the \textit{Representer theorem} \citep{steinwart2008support} tells us that the optimal solution takes the form of $f^* = \sum_{i=1}^n \alpha_i k(\cdot, x_i) = \Psi_{\bfx}\balpha$, where $\Psi_\bfx = [\psi_{x_1} \dots \psi_{x_n}]$ is the feature matrix defined by stacking feature maps along columns. If $\ell$ is the squared loss then the above optimisation is known as kernel ridge regression and $\balpha$ can be recovered in closed form $\balpha = (\bfK_{\bfx\bfx} + \lambda_fI)^{-1}\bfy$, where $\bfK_{\bfx\bfx} = \Psi_{\bfx}^\top \Psi_{\bfx}$ is the kernel matrix. If $\ell$ is the logistic loss, then the problem is known as kernel logistic regression, and $\balpha$ can be obtained using gradient descent.

\shortparagraph{Kernel embedding of distributions.} An essential component for $\textsc{RKHS-SHAP}$ is the embedding of both marginal and conditional distribution of features into the RKHS~\citep{song2013kernel, muandet2016kernel}, thus allowing one to estimate the value function analytically. Formally, the kernel mean embedding (KME) of a marginal distribution $P_X$ is defined as $\mu_X := \EE_X[\psi_X] = \int_\cX \psi_x dP_X(x)$ and the empirical estimate can be obtained as $\hat{\mu}:=\frac{1}{n}\sum_{i=1}^n \psi_{x_i}$. Furthermore, given another kernel $g:\cY \times \cY \rightarrow \RR$ with feature map $\psi_Y$ of RKHS $\cH_{g}$, the conditional mean embedding (CME) of the conditional distribution $P_{Y|X=x}$ is defined as $\mu_{Y|X=x}:= \EE[\psi_Y|X=x] = \int_{\cY} \psi_y dP_{Y|X=x}(y)$.



One way to understand CME is to view it as an evaluation of a vector-valued(VV) function $\mu_{Y\mid X}: \cX\to\cH_g$ such that $\mu_{Y\mid X}(x) = \mu_{Y\mid X=x}$, which minimises the following risk function $\EE_{p(X, Y)}[||\psi_Y - \mu_{Y\mid X}(X)||_{\cH_g}^2]$~\citep{grunewalder2012conditional}. Let $\cL(\cH_g)$ be the space of bounded linear operators from $\cH_g$ to itself. Denote $\Gamma_x:\cX\times \cX \to \cL(\cH_g)$ as the operator-valued kernel such that $\Gamma_x(x,x') = k(x, x')\mathbf{1}$ with $\mathbf{1}$ the identity operator on $\cH_g$. We denote $\cH_{\Gamma_x}$ as the corresponding vector-valued RKHS. By utilising the VV-Representer theorem~\cite{micchelli2005learning}, we could minimises the following empirical risk: $\hat{\mu}_{Y\mid X} = \underset{\mu_{Y\mid X}\in \cH_{\Gamma_x}}{\arg\min}\sum_{i=1}^n ||\psi_{y_i} - \mu_{Y\mid X}(x_i)||_{\cH_g}^2 + n\eta ||\mu_{Y\mid X}||_{\Gamma_x}^2$ where $\eta > 0$ is  regularisation parameter. This leads to the following empirical estimate of the CME, i.e., $\hat{\mu}_{Y\mid X} = \Psi_\bfy\big(\bfK_{\bfx\bfx} + n
\eta I)^{-1}\Psi_\bfx^\top$, where $\Psi_{\bfy} := \left[\psi_{y_1} ... \psi_{y_n}\right]$ and $\Psi_\bfx := \left[\psi_{x_1} ... \psi_{x_n}\right]$ are feature matrices. Intuitively, this essential turns CME estimation to a regression problem from $\cX$ to the vector-valued labels $\psi_{Y}$. Please see~\citet{micchelli2005learning} and~\citet{grunewalder2012conditional} for further discussions on vector-valued RKHSs and CMEs. In fact, when using finite-dimensional feature maps, such as in the case with running Random Fourier Features~\citep{rahimi2007random} and Nyström methods~\citep{yang2012nystrom} for scalability, one could reduce the computational complexity of evaluating empirical CME from $\mathcal{O}(n^3)$ to $\mathcal{O}(b^3) + \mathcal{O}(b^2n)$~\citep{muandet2016kernel} where $b$ is the dimension of the feature map and often can be chosen much smaller than $n$~\citep{LiTonOglSej2019}.

%% file: sections/03-main-method.tex
\vspace{-0.3cm}
\section{\textsc{RKHS-SHAP}}
\label{sec: method}
\vspace{-0.3cm}



While \textsc{KernelSHAP} is model agnostic, by restricting our attention to the class of kernel methods, faster Shapley value estimation can be derived. We assume our RKHS takes a tensor product structure, i.e, $\cH_k = \bigotimes_{i=1}^d \cH_{k^{(i)}}$, where $k^{(i)}$ is the kernel for each dimension $i\in D$. This structural assumption allows us to decompose the value functionals into tensor products of embeddings and feature maps, thus we can estimate them analytically, as later shown in Prop.~\ref{prop: representations}. Tensor product RKHSs are commonly used in practice, as they preserve universalities of kernels from individual dimension~\cite{szabo2017characteristic}, thus providing a rich function space. Note that this assumption is not essential within our framework. Namely, for a non-product kernel, one can still evaluate the value functions using tools from conditional mean embeddings and utilise our interpretability pipeline without conditional density estimation. We show this in Appendix~\ref{appendix: non-product kernels}. In the following, we will lay out the disadvantage of existing sampling and data imputation approach and show that by estimating the value functionals as elements in the RKHS, we can circumvent the need for learning and sampling from an exponential number of conditionals densities  -- thus improving the computational efficiency in the estimation.

\textbf{Estimating value functions by sampling.}\quad Estimating the Observational value function $\nu_{x, S}^{(O)}(f)$ is typically much harder than the Interventional value function $\nu_{x, S}^{(I)}(f)$ as it requires integration with respect to the unknown conditional density $p(X_{S^c}\mid X_S)$. Therefore, estimating OSVs often boils down to a two-stage approach: (1) Conditional density estimation and (2) Monte Carlo averaging over imputed data.~\citet{aas2019explaining} considered using multivariate Gaussian and Gaussian Copula for density estimation, while~\citet{frye2020shapley} proposed using deep network approaches to estimate the value function without distributional assumption. However, it is shown by \citet{yeh2022threading} recently that their approaches are not principled and generate samples that lie outside the observed data distribution. Moreover, retraining of the deep model for all possible coalitions $S$ is required, and such training is often more difficult than the training of the original model $f$.



Once the conditional density function $p(X_{S^c}\mid X_{S})$ for each $S\subseteq D$ is estimated, the observational value function at the $i^{th}$ observation $x_i$ can then be computed by taking averages of $m$ Monte Carlo samples from the estimated conditional density, i.e. $\frac{1}{m}\sum_{j=1}^{m} f(\{{x_i}_S, {x_j}_{S^c}\})$ 
where $\{{x_i}_S, {x_j}_ {S^c}\}$ is the concatenation of ${x_i}_S$ with the $j^{th}$ sample ${x_j}_{S^c}$ from $p(X_{S^c}|X_{S}={x_i}_S)$. Note further that the Monte Carlo samples cannot be reused for another observation $x_k$ as their conditional densities are different. In other words, $n \times m$ Monte Carlo samples are required for each coalition $S$ if one wishes to compute Shapley values for all $n$ observations. This is clearly not desirable. In the spirit of Vapnik's principle\footnote{\emph{When solving a problem, try to avoid solving a more general one as an intermediate step.}~\citep[Section 1.9]{vapnik95}}, as our goal is to estimate conditional expectations that lead to Shapley values, we are not going to solve a harder and more general problem of conditional density estimation as an intermediate step, but instead utilise the arsenal of kernel methods to estimate the conditional expectations directly. Further discussion on comparing complexity of \textsc{RKHS-SHAP} with density estimation methods can be found in Appendix~\ref{appendix: complexity}.

\textbf{Estimating value functions using mean embeddings.}\quad If our model $f$ lives in $\cH_k$, both the marginal and conditional expectation can be estimated analytically without any sampling or density estimation. We first show that the Riesz representations~\citep{paulsen2016introduction} of both \emph{Interventional} and \emph{Observational value functionals} exist and are well-defined in $\cH_k$. In the following, for simplicity, we will denote the functional and its corresponding Riesz representer using the same notation. For example, we will write $\nu_{x,S}(f) = \langle f, \nu_{x,S} \rangle_{\cH_k}$ when the context is clear. Given a vector of $n$ instances $\bfx$, we denote the corresponding vector of value functions as $\nu_{\bfx, S}(f) =\{\nu_{x_i, S}(f)\}_{i=1}^n$ . All proofs of this paper can be found in the Appendix~\ref{appendix: proofs}.

\begin{proposition}[Riesz representations of value functionals]
\label{prop: representations}
Denote $k$ as the product kernel of $d$ bounded kernels $k^{(i)}: \cX^{(i)} \times \cX^{(i)} \rightarrow \RR$, where $\cX^{(i)}$ is the domain of the $i^{\text{th}}$ feature for $i\in D$. Riesz representations of the Interventional and Observational value functionals then exist and can be written as     $\nu_{x,S}^{(I)} = \psi_{x_S} \otimes \mu_{X_{S^c}}$ and $\nu_{x,S}^{(O)} = \psi_{x_S} \otimes \mu_{X_{S^c}|X_S=x_{S}}$, 
where $\psi_{x_S}:=\bigotimes _{i\in S}\psi_{x^{(i)}}$, $\mu_{X_{S^c}}:=\EE[\bigotimes_{i\in S^c}\psi_{x^{(i)}}]$ and $\mu_{X_{S^c}|X_S=x_{S}} := \EE[\bigotimes_{i\in S^c}\psi_{x^{(i)}}|X_S=x_S]$.
\end{proposition}
The corresponding finite sample estimators $\hat{\nu}_{x,S}^{(I)}$ and $\hat{\nu}_{x,S}^{(O)}$ are then obtained by replacing the corresponding KME and CME components with their empirical estimators. As a result, given $f^* = \Psi_{\bfx}\balpha$ trained on dataset $(\bfx, \bfy)$, Prop.~\ref{prop: representations} allows us to estimate the value functionals analytically since $\hat{\nu}_{x, S}^{(I)}(f^*) = \langle f^*, \psi_{x_S} \otimes \hat{\mu}_{X_{S^c}} \rangle$ and $\hat{\nu}_{x, S}^{(O)}(f^*) = \langle f^*, \psi_{x_S} \otimes \hat{\mu}_{X^{S^c}|X_S=x_S} \rangle$.
This corresponds to the direct non-parametric estimators of value functions given in the following proposition, which circumvent the need for sampling or density estimation.
\begin{proposition}
\label{prop: compact-estimation}
Given $\bfx' \in \RR^{n'}$ a vector of instances and $f=\Psi_{\bfx}\balpha$, the 
empirical estimates of
the functionals can be computed as, $\hat{\nu}^{(I)}_{\bfx',S}(f) = \balpha^\top \cK_{\bfx', S}^{(I)}, \hat{\nu}^{(O)}_{\bfx',S}(f) = \balpha^\top \cK_{\bfx', S}^{(O)}$, respectively, where $\cK_{\bfx', S}^{(I)} = \bfK_{\bfx_{S}\bfx'_S} \odot \frac{1}{n}\operatorname{diag}(\bfK_{\bfx_{S^c}\bfx_{S^c}}^\top\mathbf{1_{n}}) \mathbf{1_{n}}\mathbf{1_{n'}}^\top$ and $\cK_{\bfx', S}^{(O)} = \bfK_{\bfx_S \bfx'_S}\odot \Xi_S\bfK_{\bfx_S \bfx'_S}$, $\mathbf{1_n}$ is the all-one vector with length $n$, $\odot$ the Hadamard product and $\Xi_S = \bfK_{\bfx_{S^c}\bfx_{S^c}}(\bfK_{\bfx_S\bfx_S}+n\eta I)^{-1}$.
\end{proposition}
Finally, to obtain the Shapley values with these value functions, we deploy the same least square approach as \textsc{KernelSHAP}.
\begin{proposition}[RKHS-SHAP] Given $f \in \cH_k$ and $\nu$, Shapley values $
\mathbf{B}\in\mathbb{R}^{d\times n}$ for all $d$ features and all $n$ input $\bfx$ can be computed as $\mathbf{B} = (Z^\top W Z)^{-1}Z^\top W \hat{\mathbf{V}}$ where $\hat{\mathbf{V}}_{i,:} = \hat{\nu}_{\bfx, S_i}(f)$. 
\end{proposition}

\textbf{Estimating value functions with specific models.} \quad To the best of our knowledge, TreeSHAP~\cite{lundberg2018consistent} was the only machine learning model-specific SV algorithm computing conditional expectations using the properties of the model (tree in this case) directly, rather than relying on some sort of sampling procedure and density estimation. However, it is unclear how to validate the assumptions about feature distribution in TreeSHAP, which are specified as ``the distribution generated by the tree'', as discussed by~\citet{sundararajan2020many}. In comparison, \textsc{RKHS-SHAP} does not pose assumptions on the underlying feature distribution and computes the corresponding conditional expectations via mean embeddings analytically. However, one should note that each of these model specific algorithm are only designed to explain specific models, therefore it is not informative to compare, e.g. TreeSHAP values with RKHS-SHAP values, as they are explaining different models.

\vspace{-0.2cm}
\subsection{Robustness of \textsc{RKHS-SHAP}}
\vspace{-0.1cm}
Robustness of interpretability methods is important from both an epistemic and ethical perspective, as discussed in~\citet{hancox2020robustness}. On the other hand,~\citet{alvarez2018robustness} showed empirically that Shapley methods when used with complex non-linear black-box models such as neural networks, yield explanations that vary considerably for some neighbouring inputs, even if the deep network gives similar predictions at those neighbourhoods. In light of this, we analyse the Shapley values obtained from our proposed \textsc{RKHS-SHAP} and show that they are robust. To illustrate this, we first formally define the \emph{Shapley functional},


\begin{proposition}[Shapley functional] Given a value functional $\nu$ indexed by input $x$ and coalition $S$, the Shapley functional $\phi_{x, i}:\cH_k \rightarrow \RR$ such that $\phi_{x,i}(f)$ gives the $i^{\text{th}}$ Shapley values of $x$ on $f$, has the following Riesz representation in the RKHS: $\phi_{x, i} = \frac{1}{d}\sum_{S\subseteq D\backslash\{i\}}  {d-1 \choose |S|}^{-1} \big(\nu_{x, S\cup i} - \nu_{x, S}\big)$
\vspace{-0.2cm}
\end{proposition}
Analogously, we denote $\phi_{x, i}^{(I)}$ and $\phi_{x, i}^{(O)}$ as the \emph{Interventional Shapley functional} (ISF) and \emph{Observational Shapley functional} respectively (OSF). Using the functional formalism, we now show that given $f\in\cH_k$, when $||x-x'||^2 \leq \delta$ for $\delta > 0$, the difference in Shapley values at $x$ and $x'$ will be arbitrarily small for all features i.e. $|\phi_{x,i}(f) - \phi_{x', i}(f)|$ is small $\forall i\in D$. This corresponds to the following,
\begin{align}\small
|\phi_{x,i}(f) - \phi_{x', i}(f)|^2 &= |\langle f, \phi_{x,i} - \phi_{x', i}\rangle|^2 \leq ||f||_{\cH_k}^2 ||\phi_{x,i} - \phi_{x', i}||_{\cH_k}^2
\end{align}
where we use Cauchy-Schwarz for the last line. Therefore, for a given $f$ with fix RKHS norm, the key to show robustness lies into bounding the Shapley functionals. In the following theorem, we make two assumptions: (1) the base kernels $k^{(i)}$ for each dimension $i\in D$ are bounded, and (2) the (population) conditional mean embedding functions $\mu_{X_{S^c} \mid X_{S}}$ belong to the vector-valued RKHSs $\cH_{\Gamma_{X_S}}$ for all coalitions $S\subseteq D$, therefore have finite norms. This assumption is also adopted in~\citet[Theorem 4.5]{park2020measure}.
\begin{theorem}[Bounding Shapley functionals]
Let $k$ be a product kernel with $d$ bounded kernels $|k^{(i)}(x, x)| \leq M$ for all $i\in D$. Denote $M_\mu := \sup_{S\subseteq D} M^{|S|}
, M_{\Gamma}:=\sup_{S\subseteq D} ||\mu_{X_{S^c}|X_{S}}||_{\Gamma_{X_S}}^2$ and $L_\delta = \sup_{S\subseteq D}||\psi_{x_S} - \psi_{x'_S}||_{\cH_k}^2$. Let $\delta > 0$, assume $| x^{(i)}-x^{(i)'} |^2 \leq \delta$ for all features $i\in D$, then differences of the Interventional and Observational Shapley functionals for feature $i$ at observation $x, x'$ can be bounded as $||\phi^{(I)}_{x, i} - \phi^{(I)}_{x', i}||^2_{\cH_k} \leq 2M_\mu L_\delta$ and $||\phi^{(O)}_{x, i} - \phi^{(O)}_{x', i}||^2_{\cH_k} \leq 4M_\Gamma M_\mu L_\delta$.
If $k$ is the RBF kernel with lengthscale $l$, then
\begin{align*}\small
||\phi_{x, i}^{(I)} - \phi_{x', i}^{(I)}||^2_{\cH_k} \leq 4(1 - \exp(-d\delta/2l^2)), \quad \quad ||\phi_{x, i}^{(O)} - \phi_{x', i}^{(O)}||^2_{\cH_k} \leq 8M_{\Gamma}(1 - \exp(-d\delta/2l^2))
\end{align*}


\end{theorem}
Therefore, as long as $||f||_{\cH_k}$ is small,  \textsc{RKHS-SHAP} will return robust Shapley values with respect to small perturbations. Notice the Shapley functionals do not depend on $f$ and can be estimated separately purely based on data. We will show in the next section how this key property allows us to use the functional itself to aid in learning of $f$. This enables us to enforce particular structural constraints on $f$ via an additional regularisation term.






%% file: sections/04-regulariser.tex
\vspace{-0.3cm}
\section{Shapley regularisation}
\vspace{-0.2cm}


Regularisation is popular in machine learning because it allows inductive bias to be injected to learn functions with specific properties. For example, classical $L_1$ and $L_2$ regularisers are used to control the sparsity and smoothness of model parameters. Manifold regularisation~\citep{belkin2006manifold}, on the other hand, exploits the geometry of the distribution of unlabelled data to improve learning in a semi-supervised setting, whereas~\citet{perez2017fair} and~\citet{li2019kernel} adopted a kernel dependence regulariser to learn functions for fair regression and fair dimensionality reduction. In the following, we propose a new \emph{Shapley regulariser} based on the Shapley functionals, which allows learning while controlling the level of specific feature's contributions to the model.

\textbf{Formulation}\quad Let $A$ be a specific feature whose contribution we wish to regularise, $f$ the function we wish to learn, and $\phi_{x_i, A}(f)$ the Shapley value of $A$ at a given observation $x_i$. Our goal is to penalise the mean squared magnitude of $\{\phi_{x_i, A}(f)\}_{i=1}^n$ in the ERM framework, which corresponds to $\min_{f\in\cH_{k}}\sum_{i=1}^n \ell(y_i, f(x_i)) + \lambda_f||f||_{\cH_k}^2 + \frac{\lambda_S}{n} \sum_{i=1}^n |\phi_{x_i,A}(f)|^2$, where $\ell$ is some loss function and $\lambda_f$ and $\lambda_S$ control the level of regularisations. If we replace the population Shapley functional with the finite sample estimate from Prop.~\ref{prop: representations}, and utilise the Representer theorem, we can rewrite the optimisation in terms of $\balpha$,
\begin{proposition} The above optimisation can be rewritten as, $\min_{\balpha \in \mathbb{R}^n} \sum_{i=1}^n \ell(y_i, \bfK_{x_i\bfx}\balpha) + \lambda_f \balpha^\top \bfK_{\bfx\bfx}\balpha + \frac{\lambda_S}{n} \balpha^\top \zeta_A\zeta_A^\top\balpha$.
To regularise the Interventional SVs (\textsc{ISV-Reg}) of $A$, we set $\small \zeta_A = \frac{1}{J}\sum_{j=1}^J\cK^{(I)}_{\bfx, S_j\cup A}-\cK^{(I)}_{\bfx, S_j}$ where $S_j$'s are coalitions sampled from $\small p_{SV}(S) = \frac{1}{d}{d-1 \choose |S|}^{-1}$. For regularising Observational SVs (\textsc{OSV-Reg}), we set $\small \zeta_A = \frac{1}{J}\sum_{j=1}^J\cK^{(O)}_{\bfx, S_j\cup A}-\cK^{(O)}_{\bfx, S_j}$.
\end{proposition}

In particular, closed form optimal dual weights $\balpha = (\bfK_{\bfx\bfx}^2 + \lambda_f \bfK_{\bfx\bfx} + \frac{\lambda_S}{n} \zeta_A\zeta_A^\top )^{-1}\bfK_{\bfx\bfx}\bfy$ can be recovered when $\ell$ is the squared loss. 

\shortparagraph{Choice of regularisation.} Similar to the feature attribution problem, \emph{the choice of regularising against ISVs or OSVs is application dependent} and boils down to whether one wants to take the correlation of $A$ with other features into account or not. 

\textbf{{\textsc{ISV-Reg}}}\quad \textsc{ISV-Reg} can be used to protect the model when covariate shift of variable $A$ is expected to happen at test time and one wishes to downscale $A$'s contribution during training instead of completely removing this (potentially useful) feature. Such situation may arise if, e.g., a different measurement equipment or process is used for collecting observations of $A$ during test time. \textsc{ISV} is well suited for this problem as dependencies across features will be broken by the covariate shift at test time.

\textbf{\textsc{OSV-Reg}}\quad On the other hand, \textsc{OSV-Reg} can find its application in fair learning -- learning a function that is fair with respect to some sensitive feature $A$. There exist a variety of fairness notions one could consider, such as, e.g. \emph{Statistical Parity}, \emph{Equality of Opportunity} and \emph{Equalised Odds} \citep{corbett2018measure}. In particular, we consider the fairness notion recently explored in the literature \citep{jain2020biased, mase2021cohort} that uses Shapley values, which are becoming a bridge between Explainable AI and fairness, given that they can detect biased explanations from biased models. In particular,~\citet{jain2020biased} illustrated that if a model is fair against a sensitive feature $A$, $A$ should have neither a positive nor negative contribution towards the prediction. This corresponds to $A$ having SVs with negligible magnitudes. Simply removing $A$ from the training doesn't make the model fair, as contributions of $A$ might enter the model via correlated features, therefore it is important to take feature correlations into account while regularising. Hence, it is natural to deploy \textsc{OSV-Reg} for fair learning.

%% file: sections/05-experiment-new.tex
\vspace{-0.3cm}
\section{Experiments}
\vspace{-0.2cm}


We demonstrate specific properties of \textsc{RKHS-SHAP} and Shapley regularisers using four synthetic experiments, because these properties are best illustrated under a fully controlled environment. For example, to highlight the merit of distributional-assumption-free value function estimation in RKHS-SHAP, we need groundtruth conditional expectations of value functions for verification, but they are not available in real-world data because we do not observe the true data generating distribution. Nonetheless, as model interpretability is a practical problem, we have also ran several larger scales ($n=50000, 1.8\times 10^6$) real-world explanation tasks using \textsc{RKHS-SHAP} and reported our findings in Appendix~\ref{appendix: exp} for a complete empirical demonstration. All code and implementations are made publicly available~\cite{our_repo}. 

\begin{figure*}[t!]
     \centering
     \begin{subfigure}[b]{0.66\textwidth}
         \centering
         \includegraphics[width=\textwidth]{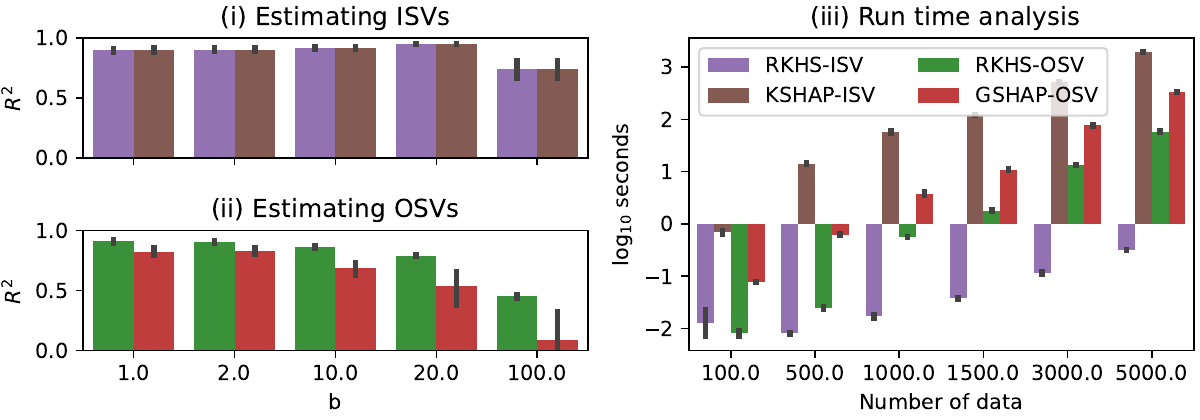}
         \caption{\small{\textsc{RKHS-SHAP} experiments}}
         \label{fig: shapley banana}
     \end{subfigure}%
     \begin{subfigure}[b]{0.33\textwidth}
         \centering
         \includegraphics[width=\textwidth]{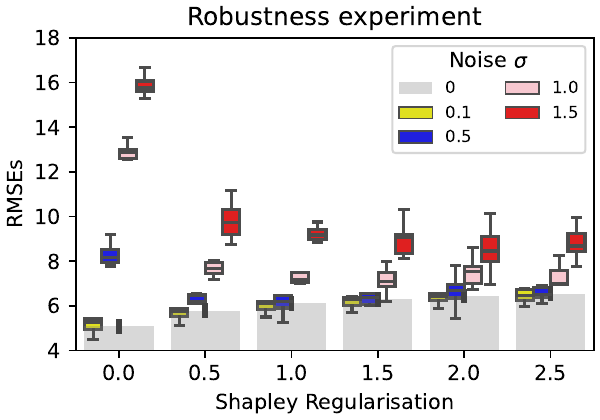}
        \caption{\small{\textsc{ISV-Reg} experiments}}
         \label{fig: ISV reg}
     \end{subfigure}%
     \vspace{-0.2cm}
        \caption{\small{(a) \textsc{RKHS-SHAP}: Estimation of Shapley values using data from the Banana distribution. Run time analysis in $\log$ scale is also reported. (b) \textsc{ISV-Reg}: RMSEs of $f_{\text{reg}}$ on noisy test data at different noise level $\sigma'$. All scores are averaged over 10 runs and 1 sd is reported.}}
        \label{fig:three graphs}
        \vspace{-0.6cm}
\end{figure*}

In the first two experiments, we evaluate \textsc{RKHS-SHAP} methods against benchmarks on estimating Interventional and Observational SVs on a Banana-shaped distribution with nonlinear dependencies~\citep{sejdinovic2014kernel}. The setup allows us to obtain closed-form expressions for the ground truth ISVs and OSVs, yet the conditional distributions among features are challenging to estimate using any standard parametric density estimation methods. We also present a run time analysis to demonstrate empirically that mean embedding based approaches are significantly more efficient than sampling based approaches. Finally, the last two experiments are applications of Shapley regularisers in robust modelling to covariate shifts and fair learning with respect to a sensitive feature.


In the following, we denote \textsc{rkhs-osv} and \textsc{rkhs-isv} as the OSV and ISV obtained from \textsc{RKHS-SHAP}. As benchmark, we implement the model agnostic sampling-based algorithm \textsc{KernelSHAP} from the Python package \textbf{shap}~\cite{lundberg2017unified}. We denote the ISV obtained from \textsc{KernelSHAP} as \textsc{kshap-isv}. As \textbf{shap} does not offer model-agnostic OSV algorithm, we implement the approach from~\citet{aas2019explaining} (described in Section 3), where OSVs are estimated using Monte Carlo samples from fitted multivariate Gaussians. We denote this approach as \textsc{gshap-osv}. We fit a kernel ridge regression on each of our experiments. Lengthscales of the kernel are selected using median heuristic~\citep{flaxman2016bayesian} and regularisation parameters are selected using cross-validation. Further implementation details and real world data illustrations are included in Appendix~\ref{appendix: exp}. 

\vspace{-0.3cm}
\subsection{\textsc{RKHS-SHAP} experiments}
\vspace{-0.2cm}
\textbf{Experiment 1: Estimating Shapley values from Banana data.} \quad We consider the following 2d-Banana distribution $\cB(b^{-1}, v)$ from~\citet{sejdinovic2014kernel}: Sample $Z\sim N(0, \operatorname{diag}(v, 1))$ and transform the data by setting $X_1 = Z_1$ and $X_2= b^{-1}(Z_1^2 - v) + Z_2$. Regression labels are obtained from $f_{\text{truth}}(X) = b^{-1}(X_1^2-v)+X_2$. This formulation allows us to compute the true ISVs and OSVs in closed forms, i.e $\phi_{X,1}^{(I)}(f_{\text{truth}}) = b^{-1}(X_1^2 - v)$, $\phi_{X,2}^{(I)}(f_{\text{truth}}) = X_2$, $\phi_{X,1}^{(O)}(f_{\text{truth}}) = \frac{1}{2}(3b^{-1}(X_1^2 - v)-X_2)$ and $\phi_{X,2}^{(O)}(f_{\text{truth}}) = \frac{1}{2}(3X_2 - b^{-1}(X_1^2 - v))$. In the following we will simulate $3000$ data points from $\cB(b^{-1}, 10)$ with $b\in [1, 10, 20, 50, 100]$, where smaller values of $b$ correspond to more nonlinearly elongated distributions. We choose $R^2$ as our metric since the true Shapley values for each experiment are scaled according to $b$. Figure \ref{fig: shapley banana}(i) and  \ref{fig: shapley banana}(ii) demonstrate $R^2$ scores of estimated ISVs and OSVs in contrast with groundtruths SVs across different configurations. We see that \textsc{rkhs-isv} and \textsc{kshap-isv} give exactly the same $R^2$ scores across configurations. This is not surprising as the two methods are mathematically equivalent. While in \textsc{kshap-isv} one averages over evaluated $\{f(x_j')\}$ with $x_j'$ being the imputed data, \textsc{rkhs-isv} aggregated feature maps of the imputed data first before evaluating at $f$, i.e $\sum_{j=1}f(x_j') = \langle f, \sum_{j=1}\phi(x_j')\rangle_{\cH_k} = \langle f, \hat{\mu}_{X}\rangle_{\cH_k}$. However, it is this subtle difference in the order of operations contribute to a significant computational speed difference as we later show in Experiment 2. In the case of estimating OSVs, we see \textsc{rkhs-osv} is consistently better than \textsc{gshap-osv} at all configurations. This highlights the merit of \textsc{rkhs-osv} as no density estimation is needed, thus avoiding any potential distribution model misspecification which happens in \textsc{gshap-osv}. 

\textbf{Experiment 2: Run time analysis.} \quad
In this experiment we sample $n$ data points from $\cB(1, 10)$ where $n\in [100, 500, 1000, 1500, 3000, 5000]$ and record the $\log_{10}$ seconds required to complete each algorithm. In practice, as the software documentation of \textbf{shap} suggests, one is encouraged to subsample their data before passing to the \textsc{KernelSHAP} algorithm as the background sampling distribution to avoid slow run time. As this approach speeds up computation at the expense of estimation accuracy since less data is used, for fair comparison with our \textsc{RKHS-SHAP} method which utilises all data, we pass the whole training set to the \textsc{KernelSHAP} algorithm. Figure \ref{fig: shapley banana}(iii) illustrates the run time across methods. We note that the difference in runtime between the two sampling based methods \textsc{kshap-isv} and \textsc{gshap-osv} can be attributed to a different software implementation, but we observe that they are both significantly slower than \textsc{rkhs-isv} and \textsc{rkhs-osv}. \textsc{rkhs-osv} is slower than \textsc{rkhs-isv} as it involves matrix inversion when computing the empirical CME. In practice, one can trivially subsample data for \textsc{RKHS-SHAP} to achieve further speedups like in the \textbf{shap} package, but one can also deploy a range of kernel approximation techniques as discussed in Section~\ref{sec: kernel methods}.

\vspace{-0.2cm}
\subsection{Shapley regularisation experiments}
\vspace{-0.1cm}

For the last two experiments we will simulate $3000$ samples from $X \sim N(0, \Sigma)$ with $\operatorname{diag}(\Sigma) = \mathbf{1_5}$ and $\Sigma_{4,5}=\Sigma_{5,4}=0.9$, 0 otherwise, therefore feature $X_4$ and $X_5$ will be highly correlated. We set our regression labels as $f_{\text{true}}(x)= x^\top \beta$ with $\beta = [1,2,3,4,10]$, enforcing $X_5$ to be the most influential feature. We use $70\%$ of our data for training and $30\%$ for testing.

\begin{wrapfigure}[20]{r}{0.27\textwidth}
  \begin{center}
  \vspace{-1cm}
    \includegraphics[width=0.27\textwidth]{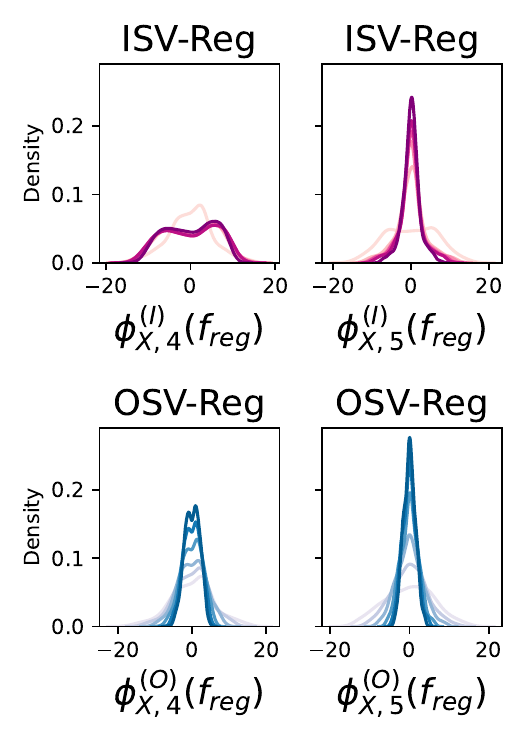}
  \end{center}
  \vspace{-0.4cm}
    \caption{\small{Distributions of SVs of sensitive feature $X_5$ and correlated feature $X_4$ obtained from $\textsc{ISV-Reg}$ and $\textsc{OSV-Reg}$ at different regularisation parameters. Colour intensity represents the strength of regularisation.}}
  \label{fig: everyone-reg}
\end{wrapfigure}

\textbf{Experiment 3: Protection against covariate shift using \textsc{ISV-Reg}.} \quad For this experiment, we inject extra mean zero Gaussian noise to the most influential feature $X_5$ in the testing set, i.e. $X_5' = X_5 + \sigma' N(0,1)$ for $\sigma' \in [0, 0.1, 0.5, 1, 1.5]$. We assume that there is an expectation for covariate shift in $X_5$ to occur at test time, due to e.g. a change in the measurement precision -- hence, we train our model $f_{\text{reg}}$ using $\textsc{ISV-Reg}$ at different regularisation level $\lambda_{s}$ for $\lambda_s \in [0, 0.5, 1, 1.5, 2, 2.5]$. We then compare RMSEs when no covariate shift is present ($\sigma' = 0)$ against RMSEs at different noise levels.  The results are shown in Figure \ref{fig: ISV reg}. We see that when no regularisation is applied, RMSEs increase rapidly as $\sigma'$ increases, indicating our standard unprotected kernel ridge regressor is sensitive to noises from $X_5'$. As the Shapley regularisation parameter increases, the RMSE of the noiseless case gradually increases too, but RMSEs of the noisy data are much closer to the noiseless case, exhibiting robustness to the covariate shift.

\textbf{Experiment 4: Fair learning with \textsc{OSV-Reg}} \quad At last, we demonstrate the use of Shapley regulariser to enable fair learning. In this context, as we will see, \textsc{OSV-Reg} is the appropriate regulariser. Consider $X_5$ as some sensitive feature which we would like to minimise its contribution during the learning of $f$. Recall $X_4$ is highly correlated to $X_5$ so it contains sensitive information from $X_5$ as well. Figure \ref{fig: everyone-reg} demonstrates how distributions of ISVs and OSVs of $X_4$ and $X_5$ changes as $\lambda_{s}$ increases. As regularisation increases, the SVs of $X_5$ becomes more centered at 0, indicating lesser contribution to the model $f_{\text{reg}}$. Similar behavior can be seen from the distribution of $\phi_{X,4}^{(O)}(f_{\text{reg}})$ but not from $\phi_{X,4}^{(I)}$. This illustrates how \textsc{ISV-Reg} will propagate unfairness through correlated feature $X_4$ while \textsc{OSV-Reg} can take them into account by minimising the contribution of sensitive information during learning.


%% file: sections/06-discussion.tex
\vspace{-0.3cm}
\section{Conclusion, limitations, and future directions}
\vspace{-0.2cm}

In this work, we proposed a more accurate and more efficient algorithm to compute Shapley values for kernel methods, termed \textsc{RKHS-SHAP}. We proved that the corresponding local attributions are robust to local perturbations under mild assumptions, a desirable property for consistent model interpretation. Furthermore, we proposed the Shapley regulariser which allows learning while controlling specific feature contribution to the model. We suggested two applications of this regulariser and concluded our work with synthetic experiments demonstrating specific aspects of our contributions. Extensive real-world data explanations are provided in Appendix~\ref{appendix: real-world} for empirical demonstration.


While our methods currently only are applicable to functions arising from kernel methods, a fruitful direction would be to extend the applicability to more general models using the same paradigm. It would also be interesting to extend our formulation to kernel-based hypothesis testing, and for example, to interpret results from two-sample tests.

%% file: sections/07-appendix.tex
›\newpage
\appendix
\onecolumn

\newtheorem{innercustomgeneric}{\customgenericname}
\providecommand{\customgenericname}{}
\newcommand{\newcustomtheorem}[2]{%
  \newenvironment{#1}[1]
  {%
   \renewcommand\customgenericname{#2}%
   \renewcommand\theinnercustomgeneric{##1}%
   \innercustomgeneric
  }
  {\endinnercustomgeneric}
}

\newcustomtheorem{customthm}{Theorem}
\newcustomtheorem{customprop}{Proposition}
\newcustomtheorem{customlemma}{Lemma}

\global\long\def\rkhs{\mathcal{H}_k}
\global\long\def\calX{\mathcal{X}}
\global\long\def\calY{\mathcal{Y}}

\section*{RKSH-SHAP: Shapley values for kernel methods supplementary materials}

\section{Computational complexity}
\label{appendix: complexity}
The gains in speed-up and accuracy in RKHS-SHAP come from estimating $\nu_{\mathbf{x},S}^{(O)}$ using Conditional Mean Embeddings (CMEs). To compare with alternative approaches, it is sufficient to look at the complexity of estimating $\nu_{\mathbf{x},S}^{(O)}(f)$. For RKHS-SHAP this is $\mathcal{O}(Nd^2m) + \mathcal{O}(N^2d^2)$ where $N$ is the number of data, $d$ is the number of Fourier features which could be taken much smaller than $N$~\citep{LiTonOglSej2019} and $m$ is the number of conjugate gradient solver steps. Previous approaches would require some form of density estimation and Monte Carlo sampling, for which there are many methods, so we present a generic decomposition of complexity here: assuming we take $L$ Monte Carlo samples for each $x_{i_{S}}$ from $p(X_{S^c}|X_S = x_{i_S})$ to estimate $\nu_{\mathbf{x}, S}^{(O)}(f)$, we have $\mathcal{O}(L^2N^2) + \mathcal{O}(\text{sampling } NL \text{ data from estimated densities}) + \mathcal{O}(\text{estimating } N \text{ conditional densities})$. It is not clear how to select $L$ nor how fast it should grow with $N$.~\citet{aas2019explaining} considered $L=N$ recovering a standard Nadaraya-Waston estimator for their empirical conditional mean estimator. In practice, for nonparametric methods, the computational cost is dominated by density estimation and sampling, both of which are not needed in our approach.

\section{RKHS-SHAP for non-product kernels}
\label{appendix: non-product kernels}
When $k$ is not a product kernel, such as the polynomial kernel and Mat\'ern kernel, we can still proceed with estimating the value function using tools from conditional mean embeddings, and utilise our interpretability pipeline without the need for solving conditional density estimation tasks. To do so, we notice that for any $f\in \cH_k$, we have
\begin{align}
    \nu_{x, S}(f) &:= \EE_X[f(X) \mid X_S=x_S] \\
                  &= \left\langle f, \EE_X[\psi_X \mid X_S=x_S]\right\rangle_{\cH_k} \\
                  &= \left\langle f, \mu_{X\mid X_S=x_S}\right\rangle_{\cH_k}.
\end{align}
Thus, we can proceed with the following estimator of $\EE_X[\psi_X \mid X_S=x_S]$ using the standard conditional mean embedding estimator (with the conditioning variable being the subset of features): Denote $k_S:\cX_S\times \cX_S \to \RR$ as a kernel defined on $\cX_S$, where $\cX_S$ is the subspace of the instance space of $\cX$ according to $S$. Note that in principle, this kernel $k_S$ need not be of the same form as the kernel $k$ defined on the full feature space, 
\begin{align}
    \hat{\mu}_{X\mid X_S=x_S} = \bfK_{x_S, X_S}\left(\bfK_{X_S, X_S} + n\lambda I\right)^{-1}\Psi_X^\top.
\end{align}
As a result, for $f=\sum_{i=1}^n \alpha_i k(\cdot,x_i)$, the corresponding non-parametric estimator of the value function $\nu_{x,S}(f)$ will be,
\begin{align}
    \bar{\nu}_{x,S}(f) = \bfK_{x_S, X_S}\left(\bfK_{X_S, X_S} + n\lambda I\right)^{-1}\bfK_{X,X}\balpha.
\end{align}
where $\balpha=(\alpha_1,\ldots,\alpha_n)^\top$.

\paragraph{Empirical demonstration} In the following, we will demonstrate the above estimation procedure to explain a kernel ridge regression learnt using Mat\'ern kernel, given by
\begin{align}
    k(x, x') =  \frac{1}{\Gamma(v)2^{v-1}}\Bigg(
\frac{\sqrt{2v}}{l} \Vert x-x' \Vert 
\Bigg)^v K_v\Bigg(
\frac{\sqrt{2v}}{l} \Vert x-x' \Vert \Bigg)
\end{align}
where $v=0.5$, $K_v$ is the modified Bessel function of the second kind, and $\Gamma$ is the gamma function. Kernel ridge regression is fitted on the diabetes and housing regression datasets from Appendix~\ref{appendix: exp}.

\begin{figure}
    \centering
    \includegraphics[width=0.49\linewidth]{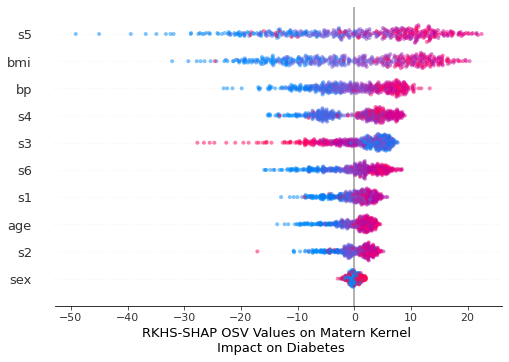}
    \includegraphics[width=0.49\linewidth]{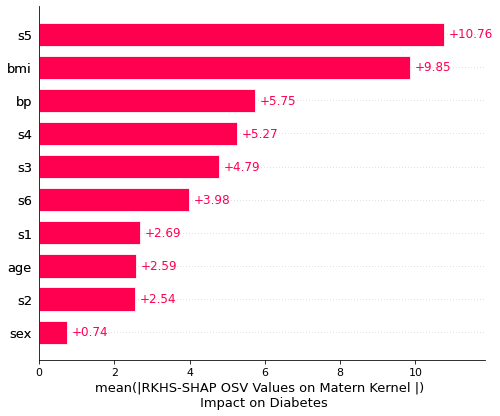}
    \caption{Explaining a Kernel Ridge regression learnt using a Mat\'ern kernel on the Diabetes regression dataset. In comparison to Figure~\ref{fig: diabetes}, where the KRR uses a Gaussian kernel, we see both models treat feature \textit{s5}, \textit{bp}, and \textit{bmi} as top predictors, but having different emphasises on features \textit{s3} and \textit{s4}.}
    \label{fig: matern diabetes}
\end{figure}

\begin{figure}
    \centering
    \includegraphics[width=0.49\linewidth]{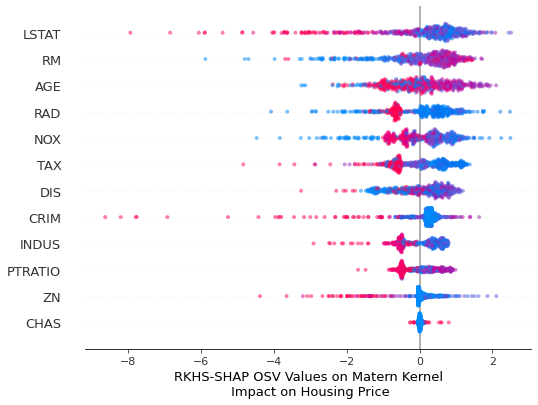}
    \includegraphics[width=0.49\linewidth]{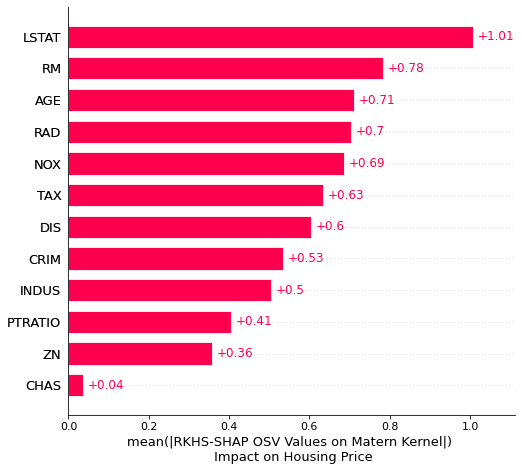}
    \caption{Explaining a Kernel Ridge regression learnt using a Mat\'ern kernel on the House price regression dataset. In comparison to Figure~\ref{fig: housing}, we see that \textit{ZN} is no longer the top predictor. This illustrated that the models emphasised the feature \textit{ZN} very differently. }
    \label{fig: matern housing}
\end{figure}

Figure~\ref{fig: matern diabetes} and \ref{fig: matern housing} illustrated the explanation results coming from the kernel ridge regression with a Mat\'ern kernel. We refer the reader to Appendix ~\ref{appendix: exp} for a guide to interpret results from the beeswarm and bar plots.

In summary, the product kernel assumption is not required for the benefits of RKHS-SHAP to be brought to bear. Our proposed framework can thus be applied to essentially any kernel appropriate for the problem at hand. It is however, required to specify the form of the said kernel for any subset of features in the case of a non-product kernel, e.g. whether it again takes a Mat\'ern form like the original kernel, or something else. Kernel hyperparameter learning will be more challenging than the product case as well, since e.g. lengthscale parameters typically vary with dimension and one would essentially require one lengthscale per subset of the features we are conditioning on, in contrast to the product case, where one lengthscale per feature dimension suffices. We might incur extra estimation error compared to the product kernel case as well. This is because one must fit the conditional mean embedding for any subset of features individually by regressing to the original RKHS defined on a higher-dimensional space (on all features $d$ rather than on the subset $|S^c|$). As an example, if $d=100$, in the non-product case we always perform estimation on the space of functions of $100$ arguments, whereas in the product case, if one is conditioning on a $|S| = 99$ dimensional subset, this simplifies to estimation on the space of functions of a scaler argument. Not only is the learning problem harder, the non-product approach has to ignore the fact that the conditioning variable here is simply the subset of features -- i.e. standard CME proceeds with regressing from features of $X_S$ to features of $X$, while in the product case it is possible to simply isolate the features we condition on, and set them to the values of interests. As a result, the product kernel assumption allows us to circumvent potential statistical errors, and thus we chose to focus on the product kernel in the main text.

\section{Proofs}
\label{appendix: proofs}
\subsection{Proof for Proposition 2.}
\begin{customprop}{2}[Riesz representations of value functionals]
Denote $k$ as the product kernel of $D$ bounded kernels $k_d: \cX^{(d)} \times \cX^{(d)} \rightarrow \RR$, where $\cX^{(d)}$ is the $d^{\text{th}}$ feature space. The Riesz representations of the Interventional value functional and Observational value functional exist and have the following forms in $\cH_k$,
\begin{align}
    \label{eq: ISF_rep}
    \nu_{x,S}^{(I)} &= \psi_{x_S} \otimes \mu_{X_{S^c}} \\
    \label{eq: OSF_rep}
    \nu_{x,S}^{(O)} &= \psi_{x_S} \otimes \mu_{X_{S^c}|X_S=x_{S}}
\end{align}
where $\psi_{x_S}:=\bigotimes _{i\in S}\psi_{x^{(i)}}$, $\mu_{X_{S^c}}:=\EE[\bigotimes_{i\in S^c}\psi_{x^{(i)}}]$ and $\mu_{X_{S^c}|X_S=x_{S}} := \EE[\bigotimes_{i\in S^c}\psi_{x^{(i)}}|X_S=x_S]$.
\end{customprop}

\begin{proof}
Since $\nu_{x,S}^{(I)}$ and $\nu_{x,S}^{(O)}$ are bounded linear functionals on all $f \in \cH_k$ with $||f||_{\cH_k}$ bounded, Riesz representation theorem \citep{paulsen2016introduction} tells us there exist $r_{\nu_{x,S}^{(I)}}$ and $r_{\nu_{x,S}^{(O)}}$ living in $\cH_k$ such that $\nu^{(I)}_{x,S}(f) = \langle f, r_{\nu_{x,S}^{(I)}} \rangle$ and $\nu^{(O)}_{x,S}(f) = \langle f, r_{\nu_{x,S}^{(O)}} \rangle$. If fact, if we set $r_{\nu_{x,S}^{(I)}}$ to be $\psi_{x_S}\otimes \mu_{X_{S^c}}$ and $r_{\nu_{x,S}^{(O)}}$ to be $\psi_{x_S}\otimes \mu_{X_{S^c}|X_S=x_S}$, then for the former, we have,
\begin{align}
    \langle f, \psi_{x_S}\otimes \mu_{X_{S^c}} \rangle &= \EE[\langle f, \psi_{x_S} \otimes \psi_{X_{S^c}}\rangle ] \\
    &= \EE[f(\{x_S, X_{S^c}\}]
\end{align}
Similarly,
\begin{align}
    \langle f, \psi_{x_S}\otimes \mu_{X_{S^c}|X_S=x_S} \rangle &= \EE[\langle f, \psi_{x_S} \otimes \psi_{X_{S^c}}\rangle |X_S=x_S] \\
    &= \EE[f(\{x_S, X_{S^c}\}|X_S=x_S]
\end{align}
\end{proof}

\subsection{Proof of Proposition 3.}

\begin{customprop}{3}
Given $\bfx' \in \RR^{n'}$ a vector of instances and $f=\Psi_{\bfx}\balpha$, the 
empirical estimates of
 $\nu^{(I)}_{\bfx',S}(f)$ and $\nu^{(O)}_{\bfx',S}(f)$ can be computed as,
 \begin{align}
     \hat{\nu}^{(I)}_{\bfx',S}(f) = \balpha^\top \cK_{\bfx', S}^{(I)} \quad\quad \hat{\nu}^{(O)}_{\bfx',S}(f) = \balpha^\top \cK_{\bfx', S}^{(O)}
 \end{align}
where $\cK_{\bfx', S}^{(I)} = \Big(\bfK_{\bfx_{S}\bfx'_S} \odot \frac{1}{n}\operatorname{diag}(\bfK_{\bfx_{S^c}\bfx'_{S^c}}^\top\mathbf{1_{n}}) \mathbf{1_{n}}\mathbf{1_{n}}^\top\Big)$ and $\cK_{\bfx', S}^{(O)} = \Big(\bfK_{\bfx_S \bfx'_S}\odot \Xi_S\bfK_{\bfx_S \bfx'_S}\Big)$, $\mathbf{1_n}$ is the all-one vector with length $n$, $\odot$ the Hadamard product and $\Xi_S = \bfK_{\bfx_{S^c}\bfx_{S^c}}(\bfK_{\bfx_S\bfx_S}+n\eta I)^{-1}$
\end{customprop}

\begin{proof} Consider $x$ a single observation. Recall $f = \Psi_{\bfx}\balpha$ and $\Psi_{\bfx} = [\psi_{x_1} ... \psi_{x_n}] = [\psi_{{x_1}_S}\otimes \psi_{{x_1}_{S^c}} ... \psi_{{x_n}_S}\otimes \psi_{{x_n}_{S^c}}]$. To compute $\hat{\nu}_{x,S}^{(I)}(f)$, we have: 
\begin{align}
    \hat{\nu}_{x,S}^{(I)}(f) &= \langle f, \psi_{x_S}\otimes \hat{\mu}_{X_{S^c}}\rangle \\
                             &= \langle \Psi_{\bfx}\balpha, \psi_{x_S} \otimes \frac{1}{n}\sum_{i=1}^n \psi_{{x_i}_{S^c}}\rangle \\
                             &= \balpha ^\top \big(\bfK_{\bfx_{S} x_S} \times \frac{1}{n}\bfK_{\bfx_{S^c} x_{S^c}}^\top {\bf1}_n \big)
\end{align}                             
Similarly, for $\nu_{x, S}^{(O)}(f)$,
\begin{align}
\hat{\nu}_{x, S}^{(O)}(f) &= \langle f,\psi_{x_S}\otimes \hat{\mu}_{X_{S^c}|X_S=x_S} \rangle \\
                              &= \langle \Psi_{\bfx}\balpha, \psi_{x_S} \otimes \Psi_{\bfx_{S^c}}(\bfK_{{\bfx}_S{\bfx}_S}+\eta I)^{-1}\bfK_{\bfx_Sx_S}\rangle\\
                              &= \balpha^\top\big(\bfK_{\bfx_Sx_S}\odot \bfK_{\bfx_{S^c}\bfx_{S^c}}(\bfK_{{\bfx}_S{\bfx}_S}+n\eta I)^{-1}\bfK_{\bfx_Sx_S}\big)
\end{align}
Extension to a vector of instance $\bfx'$ is then straight forward.
\end{proof}

\subsection{Proof of Proposition 4.}

\begin{customprop}{4}[RKHS-SHAP] Given $f \in \cH_k$ and a value functional $\nu$, Shapley values for all $d$ features and all input $\bfx$ can be computed as follows:
\begin{align}
    \mathbf{B} = (Z^\top W Z)^{-1}Z^\top W \hat{\mathbf{V}}
\end{align}
where $\hat{\mathbf{V}}_{i,:} = \langle f, \hat{\nu}_{\bfx, S_i}\rangle$.
\end{customprop}

\paragraph{Proof} Since we now have a compact way to estimate the conditional estimations for a vector of observations using mean embeddings, we can restate the KernelSHAP objective, which essentially is a weighted least regression, into a multi-output weighted least square formulation.

\subsection{Proof of Proposition 5.}

\begin{customprop}{5}[Shapley functional] Given a value functional $\nu$ indexed by input $x$ and coalition $S$, the Shapley functional $\phi_{x, i}:\cH_k \rightarrow \RR$ such that $\phi_{x,i}(f)$ is the $i^{\text{th}}$ Shapley values for model $f$ on input $x$, has the following Reisz representation in the RKHS,
\begin{align}
    \phi_{x, i} = \frac{1}{d}\sum_{S\subseteq D\backslash\{i\}}  {d-1 \choose |S|}^{-1} \big(\nu_{x, S\cup i} - \nu_{x, S}\big)
\end{align}
\end{customprop}

\paragraph{Proof} Since the Shapley functional is a linear combination of bounded linear functionals (value functionals), it admits a Riesz representer in the RKHS.

\subsection{Proof of Theorem 6.}

\begin{customthm}{6}[Bounding Shapley functionals]
Let $k$ be a product kernel with $d$ bounded kernels $|k^{(i)}(x, x)| \leq M$ for all $i\in D$. Denote $M_\mu := \sup_{S\subseteq D} M^{|S|}
, M_{\Gamma}:=\sup_{S\subseteq D} ||\mu_{X_{S^c}|X_{S}}||_{\Gamma_{X_S}}^2$ and $L_\delta = \sup_{S\subseteq D}||\psi_{x_S} - \psi_{x'_S}||_{\cH_k}^2$. Let $\delta > 0$, assume $|x^{(i)}-x^{(i)'}|^2 \leq \delta$ for all features $i\in D$, then differences of the Interventional and Observational Shapley functionals for feature $i$ at observation $x, x'$ can be bounded as $||\phi^{(I)}_{x, i} - \phi^{(I)}_{x', i}||^2_{\cH_k} \leq 2M_\mu L_\delta$ and $||\phi^{(O)}_{x, i} - \phi^{(O)}_{x', i}||^2_{\cH_k} \leq 4M_\Gamma M_\mu L_\delta$.
If $k$ is the RBF kernel with lengthscale $l$, then \begin{align}
    \small
    ||\phi_{x, i}^{(I)} - \phi_{x', i}^{(I)}||^2_{\cH_k} &\leq 4\Big(1 - \exp\Big(\frac{-d\delta}{2l^2}\Big)\Big) \\
    ||\phi_{x, i}^{(O)} - \phi_{x', i}^{(O)}||^2_{\cH_k} &\leq 8M_{\Gamma}\Big(1 - \exp\Big(\frac{-d\delta}{2l^2}\Big)\Big)
\end{align}
\end{customthm}


\paragraph{Proof } To prove that Shapley functionals between two observations $x$ and $x'$ are $\delta$ close when the two points are closed, we proceed as follows: (1) We show that when one pick the usual product RBF kernel, we can bound the distance of the feature maps as a function of $\delta$. (2) We then upper bound the value functionals and show that this bound can be relaxed so that it is independent with the choice of coalition. (3) Since Shapley values is an expectation of differences of value functions, by devising a coalition independent bound for the difference in value functionals, the expectation disappears in our bound. 

\begin{customprop}{A.1}[Bounding feature maps] For the simplest 1 dimensional case with $|x-x'|^2 \leq \delta$, if we pick $k$ the standard RBF kernel with lengthscale $l$, we have,
\begin{align}
    ||\psi_x - \psi_{x'}||_{\cH_k}^2 \leq 2 - 2\exp\Big(-\frac{\delta}{2l^2}\Big)
\end{align}
When we pick $x, x' \in \RR^{d}$ and with a product RBF kernel i.e $k(x, x') = \prod_{j=1}^d k^{j}(x^{(j)}, x'^{(j)})$, where $k^{(j)}$ themselves RBF kernels. For simplicity, we assume they all share the same lengthscale $l$. If $|x^{(j)} - x'^{(j)}| \leq \delta$ for all $j \in D$, then we can bound the difference in feature maps as follows,
\begin{align}
    ||\psi_x - \psi_{x'}||_{\cH_k}^2 \leq 2 - 2\exp\Big(-\frac{d\delta}{2l^2}\Big)
\end{align}
\end{customprop}

\begin{proof}
Since $||\psi_x - \psi_{x'}||^2_{\cH_k} = k(x, x) + k(x, x') - 2k(x, x')$. Therefore the first 2 terms are $1$ and we can bound the last term since,
\begin{align}
    k(x, x') = \exp\Big(-\frac{|x-x'|^2}{2l^2}\Big) \geq \exp\Big(-\frac{\delta}{2l^2}\Big)
\end{align}
Multiply this lower bound $d$ times to obtain the bound for the $d$ dimensional case.
\end{proof}

Proposition A.1 tells us how the distance in feature maps $||k_x - k_{x'}||_{\cH_k}$ can be expressed by the distance between $x$ and $x'$ in the RBF kernel. Different bounds can be derived for different kernels and we only show the special RBF case for illustration purpose.

Now we shall prove a bound for the value functionals. We shall first proceed with the interventional case and move on to observational afterwards.

\begin{customprop}{A.2}[Bounding Interventional value functionals]
For a fix coalition $S$, denote $D_S^{(I)} = ||\nu_{x, S}^{(I)} - \nu_{x', S}^{(I)}||_{\cH_k}^2$. Then $D_S^{(I)} \leq ||\psi_{x_S} - \psi_{x'_S}||^2_{\cH_{k_{S}}}||\mu_{X_{S^c}}||_{\cH_{k_{S^c}}}^2$. Let $L_\delta:= \sup_{S \subseteq D}||\psi_{x_S} - \psi_{x'_S}||^2_{\cH_{k_{S}}}$ and assume kernels are all bounded per dimension by $M$, i.e $k^{(j)}(x, x') \leq M$ for all $j \in D$. Denote $M_\mu:=\sup_{S \subseteq D}M^{|S|}$. Then the bound can be further loosen up,
\begin{align}
    D_S^{(I)} \leq M_{\mu}L_\delta
\end{align}
\end{customprop}
\begin{proof}
\begin{align}
    D_S^{(I)}
    &=||\nu_{x, S}^{(I)} - \nu_{x', S}^{(I)}||_{\cH_k}^2 \\ 
    &= ||\psi_{x_S} \otimes \mu_{X_{S^c}} - \psi_{x'_S} \otimes \mu_{X_{S^c}}||_{\cH_k}^2 \\
    & \leq||\psi_{x_S} - \psi_{x'_S}||^2_{\cH_{k_{S}}}||\mu_{X_{S^c}}||^2_{\cH_{k_{S^c}}}
\intertext{Note that $||\mu_{X_{S^c}}||^2 = ||\EE[k(X_{S^c}, X'_{S^c})]||^2 \leq M^{2|S^c|}$, therefore,}
    &\leq L_\delta M_\mu
\end{align}
\end{proof}

Before we proof the main theorem, we will illustrate the following bounds for conditional mean embeddings, which will be used to bound the observational shapley functionals.

\begin{customprop}{A.3}[Bounding conditional mean embeddings] If we take on the vector-valued function perspective of conditional mean embeddings as in~\citet{grunewalder2012conditional}, then we could assume, in general for random variable $Y$ and $X$, there exits a function $\mu_{Y|X} \in \cH_{\Gamma_{x}}$ where $\Gamma_x: \cX \times \cX \mapsto \cL(\cH_{\ell})$ with $\cL(\cH_{\ell})$ being the space of self-adjoint operators from the RKHS $\cH_\ell$ to itself, is the vector-valued kernel $\Gamma_x(x,x') = k(x,x') \mathbf{1}$, such that $\mu_{Y|X}(x) = \mu_{Y\mid X=x}$. If we assume such function exists, then by definition of vector-valued RKHSs as in~\citet{park2020measure}, $||\mu_{Y\mid X}||_{\cH_{\Gamma_x}}$ has finite norm. Therefore the following is defined if the base kernel $k$ is bounded,
\begin{align}
    ||\mu_{Y\mid X=x}||_{\cH_\ell}
    & \leq ||\mu_{Y\mid X}||_{\cH_{\Gamma_x}} ||\psi_x||_{\cH_k}
\end{align}

and correspondingly, 
\begin{align}
    ||\mu_{Y\mid X=x} - \mu_{Y\mid X=x'}||_{\cH_{\ell}} \leq ||\mu_{Y\mid X}||_{\cH_{\Gamma_x}} ||\psi_x - \psi_{x'} ||
\end{align}
\end{customprop}
\begin{proof}
For the first claim, using the result from~\citet[Prop.1 ]{micchelli2005learning}, we have,
\begin{align}
||\mu_{Y\mid X}(x)||_{\cH_\ell} \leq ||\mu_{Y|X}||_{\cH_{\Gamma_x}} ||\Gamma_x(x,x)||_{op}^{\frac{1}{2}}
\end{align}
however, we have,
\begin{align}
    ||\Gamma_x(x,x)||_{op} = \sup_{g\in \cH_{\ell}} \frac{||k(x,x)g||_{\cH_{\ell}}}{||g||_{\cH_{\ell}}} = |k(x,x)| = ||\psi_x||^2
\end{align}
For the second part, we start with,
\begin{align}
    ||\mu_{Y\mid X=x} - \mu_{Y\mid X=x'}||_{\cH_{\ell}} \leq ||\mu_{Y\mid X}||_{\cH_{\Gamma_x}} ||\Gamma_x(\cdot, x) - \Gamma_x(\cdot, x')||_{op}
\end{align}
Using the result from~\citet[Prop.1]{micchelli2005learning} again, we have
\begin{align}
    ||\Gamma_x(\cdot, x) - \Gamma_x(\cdot, x')||_{op} &= ||\big(\Gamma_x(\cdot, x) - \Gamma_x(\cdot, x')\big)^*\big(\Gamma_x(\cdot, x) - \Gamma_x(\cdot, x')\big)||_{op}^\frac{1}{2} \\
\intertext{where the $*$ denotes the adjoint of the operator,}
    &= ||\Gamma_x(\cdot, x)^*\Gamma_x(\cdot, x) - 2 \Gamma_x(\cdot, x)^*\Gamma_x(\cdot, x') + \Gamma_x(\cdot, x')^*\Gamma_x(\cdot, x')||_{op}^{\frac{1}{2}} \\
    &= ||\Gamma_x(x,x) -2 \Gamma_x(x, x') + \Gamma_x(x',x')||_{op}^\frac{1}{2} \\
    &= ||\big(k(x,x) - 2k(x,x') + k(x',x')\big) \mathbf{1}||_{op}^{\frac{1}{2}} \\
    &= ||\psi_x - \psi_{x'}||_{\cH_k}
\end{align}
therefore we have as a result,
\begin{align}
    ||\mu_{Y\mid X=x} - \mu_{Y\mid X=x'}||_{\cH_{\ell}} \leq ||\mu_{Y\mid X}||_{\cH_{\Gamma_x}} ||\psi_x - \psi_{x'}||_{\cH_k}
\end{align}
\end{proof}

\begin{customprop}{A.4}[Bounding Observational value functionals via vector-valued function perspective of CME]
For a fix coalition $S$, denote $D_S^{(O)} = ||\nu_{x, S}^{(I)} - \nu_{x', S}^{(I)}||_{\cH_k}^2$. Then $D_S^{(O)} \leq ||\psi_{x_S} - \psi_{x'_S}||^2_{\cH_{k_S}}||\mu_{X_{S^c}|X_S}||_{\cH_{\Gamma_{X_S}}}^2\big(||\psi_{x_S}||_{\cH_{k_S}}^2 + ||\psi_{x'_S}||_{\cH_{k_S}}^2\big)$, where $\cH_{\Gamma_{X_S}}$ is the $\cH_{k_{S^c}}$-valued RKHS. If we denote $L_\delta = \sup_{S \subseteq D}||\psi_{x_S} - \psi_{x'_S}||_{\cH_{k_S}}^2$ and $M_{\mu} = \sup_{S\subseteq D}M^{|S|}$ and $M_\Gamma = \sup_{S\subseteq D} ||\mu_{X_{S^c}|X_S}||_{\cH_{\Gamma_{X_S}}}^2$. Then $D_S^{(O)} \leq 2 M_\Gamma M_\mu L_\delta$ for all coalition $S$. 
\end{customprop}

\begin{proof}
\begin{align}
    D_S^{(O)} &= ||\nu_{x, S}^{(O)} - \nu_{x', S}^{(O)}||_{\cH_{k}}^2 \\
              &= ||\psi_{x_S}\otimes \mu_{X_{S^c}|X_{S}=x_S} - \psi_{x'_S}\otimes \mu_{X_{S^c}|X_{S}=x'_S}||_{\cH_k}^2 \\
              &=||\psi_{x_S}\otimes \mu_{X_{S^c}|X_{S}=x_S} -\psi_{x'_S}\otimes \mu_{X_{S^c}|X_{S}=x_S} + \psi_{x'_S}\otimes \mu_{X_{S^c}|X_{S}=x_S}-  \psi_{x'_S}\otimes \mu_{X_{S^c}|X_{S}=x'_S}||_{\cH_k}^2 \\
              &\leq ||\psi_{x_S} - \psi_{x'_S}||^2_{\cH_{k_S}}||\mu_{X_{S^c}|X_S=x_s}||_{\cH_{k_{S^c}}}^2 + ||\psi_{x'_S}||^2_{\cH_{k_S}}||\mu_{X_{S^c}|X_S=x_S} - \mu_{X_{S^c}|X_S=x'_S}||_{\cH_{k_{S^c}}}^2 \\
              &\leq ||\psi_{x_S} - \psi_{x'_S}||^2_{\cH_{k_S}}||\mu_{X_{S^c}|X_S}||_{\cH_{\Gamma_{X_S}}}^2||\psi_{x_S}||_{\cH_{k_S}}^2 + ||\psi_{x'_S}||^2_{\cH_{k_S}}||\mu_{X_{S^c}|X_S}||_{\cH_{\Gamma_{X_S}}}^2||\psi_{x_S} - \psi_{x'_S}||_{\cH_{k_S}}^2 \\
              &= ||\psi_{x_S} - \psi_{x'_S}||^2_{\cH_{k_S}}||\mu_{X_{S^c}|X_S}||_{\cH_{\Gamma_{X_S}}}^2\big(||\psi_{x_S}||_{\cH_{k_S}}^2 + ||\psi_{x'_S}||_{\cH_{k_S}}^2\big) \\
              &\leq 2M_\Gamma M_\mu L_\delta
\end{align}
Finally, we note that,
\begin{align}
    ||\phi_{x,i} - \phi_{x', i}||_{\cH_{k}}^2 &= ||\frac{1}{d}\sum_{S \subseteq D\backslash{\{i\}}}{d-1 \choose |S|}^{-1}\nu_{x, S\cup i} - \nu_{x, S} - (\nu_{x', S\cup i} - \nu_{x', S})||_{\cH_{k}}^2 \\
    &\leq \frac{1}{d}\sum_{S \subseteq D\backslash{\{i\}}}{d-1 \choose |S|}^{-1} D_S + D_{S \cup i} \\
    &= \EE_S[D_S + D_{S\cup i}]
\end{align}
Since we have proven bounds for $D_S^{(O)}$ and $D_S^{(I)}$ that is coalition independent, we can directly substitute the bound inside the expectation. Therefore
\begin{align}
    ||\phi^{(I)}_{x,i} - \phi^{(I)}_{x', i}||_{\cH_{k}}^2 &\leq 2L_\delta M_\mu \\
    ||\phi^{(O)}_{x,i} - \phi^{(O)}_{x', i}||_{\cH_{k}}^2 &\leq 4M_\Gamma L_\delta M_\mu
\end{align}
In the case when we pick $k$ as a product RBF kernel, we have $L_\delta = 2-2\exp\Big(\frac{d\delta}{2l^2}\Big)$ and $M_\mu = 1$, therefore,
\begin{align}
    ||\phi_{x, i}^{(I)} - \phi_{x', i}^{(I)}||^2_{\cH_k} &\leq 4\Bigg(1 - \exp\bigg(\frac{-d\delta}{2l^2}\bigg)\Bigg) \\
    ||\phi_{x, i}^{(O)} - \phi_{x', i}^{(O)}||^2_{\cH_k} &\leq 8M_{\Gamma}\Bigg(1 - \exp\bigg(\frac{-d\delta}{2l^2}\bigg)\Bigg)
\end{align}
\end{proof}

\begin{customprop}{7} The above optimisation can be rewritten as, $\min_{\balpha \in \mathbb{R}^n} \sum_{i=1}^n \ell(y_i, \bfK_{x_i\bfx}\balpha) + \lambda_f \balpha^\top \bfK_{\bfx\bfx}\balpha + \frac{\lambda_S}{n} \balpha^\top \zeta_A\zeta_A^\top\balpha$.
To regularise the Interventional SVs (\textsc{ISV-Reg}) of $A$, we set $\small \zeta_A = \frac{1}{J}\sum_{j=1}^J\cK^{(I)}_{\bfx, S_j\cup A}-\cK^{(I)}_{\bfx, S_j}$ where $S_j$'s are coalitions sampled from $\small p_{SV}(S) = \frac{1}{d}{d-1 \choose |S|}^{-1}$. For regularising Observational SVs (\textsc{OSV-Reg}), we set $\small \zeta_A = \frac{1}{J}\sum_{j=1}^J\cK^{(O)}_{\bfx, S_j\cup A}-\cK^{(O)}_{\bfx, S_j}$.
\end{customprop}

\begin{proof}[Sketch proof.]
To express $$\min_{f\in\cH_{k}}\sum_{i=1}^n \ell(y_i, f(x_i)) + \lambda_f||f||_{\cH_k}^2 + \frac{\lambda_S}{n} \sum_{i=1}^n |\phi_{x_i,A}(f)|^2$$ as $$\min_{\balpha \in \mathbb{R}^n} \sum_{i=1}^n \ell(y_i, \bfK_{x_i\bfx}\balpha) + \lambda_f \balpha^\top \bfK_{\bfx\bfx}\balpha + \frac{\lambda_S}{n} \balpha^\top \zeta_A\zeta_A^\top\balpha,$$ it suffices to show that $\frac{\lambda_S}{n} \sum_{i=1}^n |\phi_{x_i,A}(f)|^2 =  \frac{\lambda_S}{n} \balpha^\top \zeta_A\zeta_A^\top\balpha$. However, note that
\begin{align}
    \sum_{i=1}^n |\phi_{x_i,A}(f)|^2 &= \phi_{\bfx,A}(f)^\top \phi_{\bfx,A}(f) \\
    &= f^\top \phi_{\bfx,A} \phi_{\bfx,A}^\top f
\end{align}
Now we can estimate the Shapley functional $\phi_{\bfx,A}$ defined in Proposition 5, by applying the finite sample estimator of the value functions from Proposition 2, we can compute the finite sample estimate of $\phi_{\bfx,A}^\top f$ as $\zeta_A^\top \balpha$.
\end{proof}

\section{Further experiment details}
\label{appendix: exp}
\subsection{Banana Distribution $\mathcal{B}(b^{-1}, v)$}

Recall the Banana distribution $\mathcal{B}(b^{-1},v)$ is defined as follows: Let $Z \sim N(0, \operatorname{diag}(v, 1)) $and set $X_1 = Z_1$ and $X_2 = b^{-1}(Z_1^2 - v) + Z_2$. We define $f(x) = b^{-1}(x_1^2 - v) + x_2$. Now then we have,
\begin{align}
    \EE[f(X)] &= 0 \\
    \EE[f(X)|X_1 = x_1] &= 2b^{-1}(x_1^2 - v) \\
    \EE[f(X)|X_2 = x_2] &= 2x_2 \\
    \EE[f(X)|do(X_1) = x_1] &= b^{-1}(x_1^2 - v)\\
    \EE[f(X)|do(X_2) = x_2] &= x_2
\end{align}

This corresponds to the following Observational Shapley values,
\begin{eqnarray*}
\phi_{x,1}^{(O)}(f) & = & \frac{1}{2}\left[\binom{1}{0}^{-1}\left(\mathbb{E}f\left({\bf X}|X_{1}=x_{1}\right)-\mathbb{E}f\left({\bf X}\right)\right)+\binom{1}{1}^{-1}\left(\mathbb{E}f\left({\bf X}|X_{1}=x_{1},X_{2}=x_{2}\right)-\mathbb{E}f\left({\bf X}|X_{2}=x_{2}\right)\right)\right]\\
 & = & \frac{1}{2}\left(3b^{-1}\left(x_{1}^{2}-v\right)-x_{2}\right).\\
\phi_{x,2}^{(O)}(f) & = & \frac{1}{2}\left[\binom{1}{0}^{-1}\left(\mathbb{E}f\left({\bf X}|X_{2}=x_{2}\right)-\mathbb{E}f\left({\bf X}\right)\right)+\binom{1}{1}^{-1}\left(\mathbb{E}f\left({\bf X}|X_{1}=x_{1},X_{2}=x_{2}\right)-\mathbb{E}f\left({\bf X}|X_{1}=x_{1}\right)\right)\right]\\
 & = & \frac{1}{2}\left(3x_{2}-b^{-1}\left(x_{1}^{2}-v\right)\right)
\end{eqnarray*}

Similarly, for Interventional Shapley values we have,
\begin{align*}
    \phi_{x,1}^{(I)}(f) &= b^{-1}(x_1^2 - v) \\
    \phi_{x, 2}^{(I)}(f) &= x_2
\end{align*}

\subsection{\textsc{RKHS-SHAP} on real-world examples}
\label{appendix: real-world}

We demonstrate the result of running \textsc{RKHS-SHAP} on 6 real-world datasets and showcase their RKHS-SHAP Observational Shapley values in Beeswarm summary plots. Interventional SVs are omitted because we have shown in the main text that running KernelSHAP-ISV and RKHS-SHAP-ISV gives you the same SVs, and they only differ in computational run time.

These results are not included in the main text because \textbf{we do not observe the actual data distribution, thus there are no groundtruth observational SVs that our algorithm can be compared to measure and verify how well it is performing}. In the following, all models are fitted with the Gaussian kernel. We first fit a Kernel Ridge Regression or Kernel Logistic Regression to learn the function $f$, and apply \textsc{RKHS-SHAP} to $f$ to recover the corresponding observational Shapley values.

We present our results using Beeswarm plot and bar plot. According to the \textbf{shap} package, the beeswarm plot is designed to display an information-dense summary of how the top features in the dataset impact the model's output. Each instance the given explanation is represented by a single dot on each feature row. The x position of the dot is determined by the RKHS-SHAP value of that feature, and dots “pile up” along each feature row to show density. Colour is used to display the original value of a feature, which is scaled with red indicating high, and blue indicating low values. On the other hand, the bar plot shows the mean absolute value of the Shapley values per feature, thus providing some global summary based on recovered local importances.

We summarise our real-world explanation tasks in table~\ref{table: real-world}.

\begin{table}[]
\caption{Real-world explanation tasks}
\centering
\begin{tabular}{|c|c|c|c|}
\hline
Dataset                                                                         & $n_{instances}$ & $n_{features}$ & Downstream task                                                                                                       \\ \hline\hline
\begin{tabular}[c]{@{}c@{}}Boston \\ Housing\end{tabular}                       & 506              & 12              & \begin{tabular}[c]{@{}c@{}}Predict Boston House Price \\ (Regression)\end{tabular}                                    \\ \hline
\begin{tabular}[c]{@{}c@{}}Diabetes \\ Progression\end{tabular}                 & 442              & 10              & \begin{tabular}[c]{@{}c@{}}Predict diabetes progression\\ (Regression)\end{tabular}                                   \\ \hline
\begin{tabular}[c]{@{}c@{}}Diabetes for \\ Pima Indian \\ Heritage\end{tabular} & 768              & 8               & \begin{tabular}[c]{@{}c@{}}Predict whether a patient has diabetes\\ (Classification)\end{tabular}                     \\ \hline
Breast Cancer                                                                   & 569              & 30              & \begin{tabular}[c]{@{}c@{}}Predict whether a patient might have \\ breast cancer or not (Classification)\end{tabular} \\ \hline
Census Income                                                                   & 48,842           & 14              & \begin{tabular}[c]{@{}c@{}}Predict whether an individual is making \\ over \$50k a year (Classification)\end{tabular} \\ \hline
\begin{tabular}[c]{@{}c@{}}League of Legends\\ Win Prediction\end{tabular}      & 1,800,000          & 71              & \begin{tabular}[c]{@{}c@{}}Predict the winning probability of a player\\ (Classification)\end{tabular}                 \\ \hline
\end{tabular}
\label{table: real-world}
\end{table}

\paragraph{Boston Housing} The Boston house price  dataset\footnote{https://archive.ics.uci.edu/ml/machine-learning-databases/housing/} contains $506$ instances and $12$ numerical features. Below is the description of its features:
\begin{itemize}
    \item[] 
    \begin{itemize}
    \item[CRIM] per capita crime rate by town
    \item[ZN] proportion of residential land zoned for lots over 25k sq.ft
    \item[INDUS] proportion of non-retail business acres per town
        \item[CHAS]     Charles River dummy variable (= 1 if tract bounds river; 0 otherwise)
        \item[NOX]      nitric oxides concentration (parts per 10 million)
        \item[RM]       average number of rooms per dwelling
        \item[AGE]      proportion of owner-occupied units built prior to 1940
        \item[DIS]      weighted distances to five Boston employment centres
        \item[RAD]      index of accessibility to radial highways
        \item[TAX]      full-value property-tax rate per \$10,000
        \item[PTRATIO]  pupil-teacher ratio by town
        \item[LSTAT]    \% lower status of the population
        \item[MEDV]     Median value of owner-occupied homes in \$1000's    
\end{itemize}
\end{itemize}

\begin{figure}[h!]
    \centering
    \includegraphics[width=0.55\textwidth]{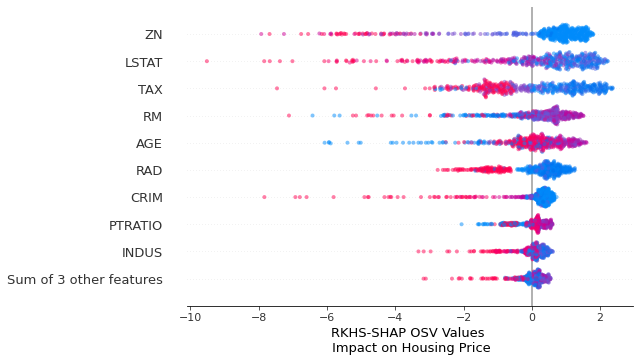}
    \includegraphics[width=0.44\textwidth]{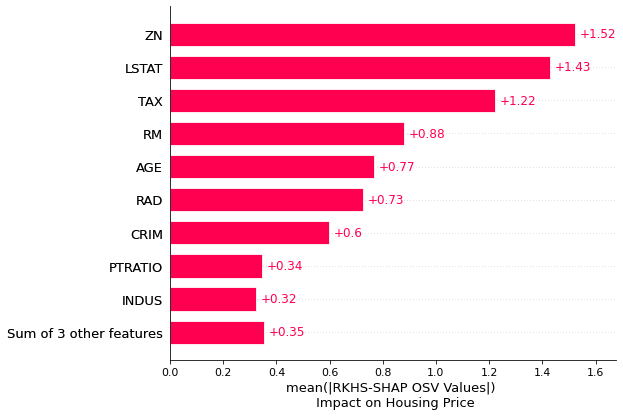}
    \caption{Beeswarm and bar plot for the housing dataset.}
    \label{fig: housing}
\end{figure}

We fit a Kernel Ridge Regression to predict the Boston house price. The results are shown in Fig.~\ref{fig: housing}. We see that RKHS-SHAP does capture several intuitive explanations, e.g. Higher crime rate (red dots in feature CRIM) corresponds to negative impact on the house price. We also recover explanations such as lower percentage of lower status of the population (LSTAT) will increase the house price.

\begin{figure}[h!]
    \centering
    \includegraphics[width=0.49\textwidth]{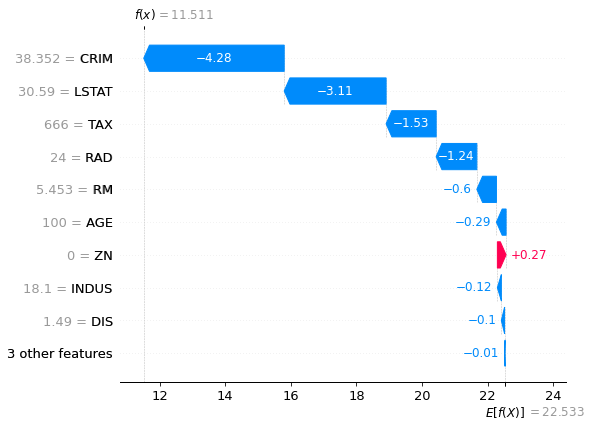}
    \includegraphics[width=0.49\textwidth]{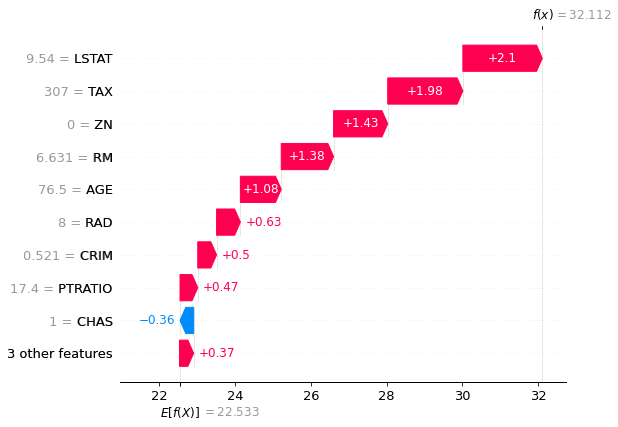}
    \caption{(left) The algorithm believes having a high crime rate is the major reason for its low house price. (Right) Having a high LSTAT increased the house price.}
    \label{fig: individual house}
\end{figure}

We can examine specific houses and interpret why the kernel ridge regression predicts their corresponding house prices as well. See Fig~\ref{fig: individual house}.

\begin{figure}[h!]
    \centering
    \includegraphics[width=0.55\textwidth]{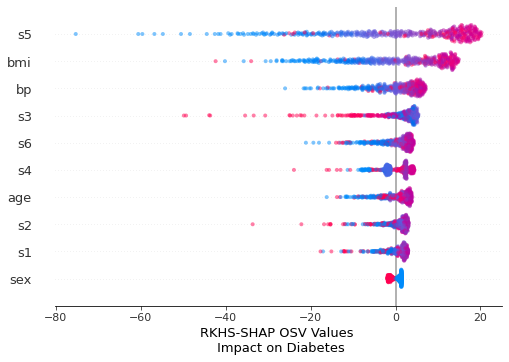}
    \includegraphics[width=0.44\textwidth]{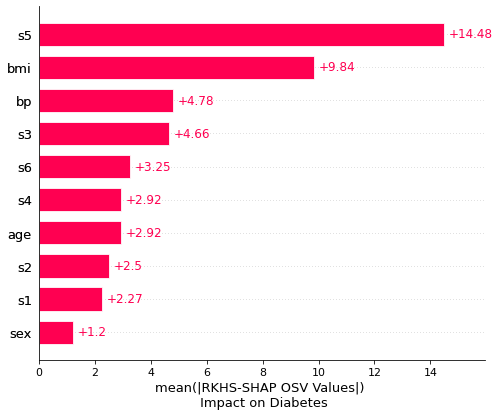}
    \caption{Beeswarm and barplot of the RKHS-SHAP values on the Diabetes dataset}
    \label{fig: diabetes}
\end{figure}

\paragraph{Diabetes progression regression} Next we apply \textsc{RKHS-SHAP} to the diabetes\footnote{https://www4.stat.ncsu.edu/~boos/var.select/diabetes.html} dataset with $442$ samples and $10$ features. The machine learning task is to model the disease progression of patients as a regression problem. We fit a kernel ridge regression for that.  Figure \ref{fig: diabetes} records the results. Feature $s1$ to $s6$ are blood serum measurements. We note that $bmi$ is one of the most influential feature, which follows our intuition that higher value of $bmi$ (red clusters in the bmi row) should be a strongly predictive variable to diabetes.

\paragraph{Diabetes for Pima Indian heritage} Here we consider another dataset of diabetes study for Pima Indian heritage women aged 21 over. The data set is collected from here\footnote{https://www.kaggle.com/datasets/mathchi/diabetes-data-set?resource=download}. There are 768 samples with 8 features. The goal is to predict whether a patient has diabetes and fit a kernel logistic regression.

\begin{figure}[h!]
    \centering
    \includegraphics[width=0.55\textwidth]{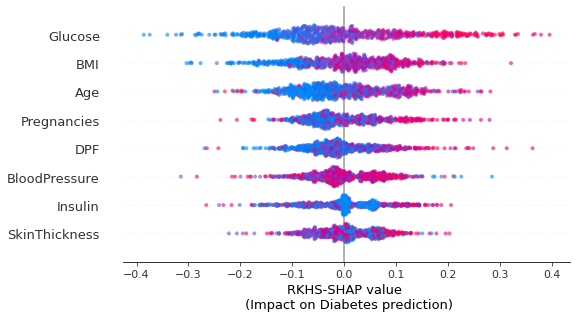}
    \includegraphics[width=0.44\textwidth]{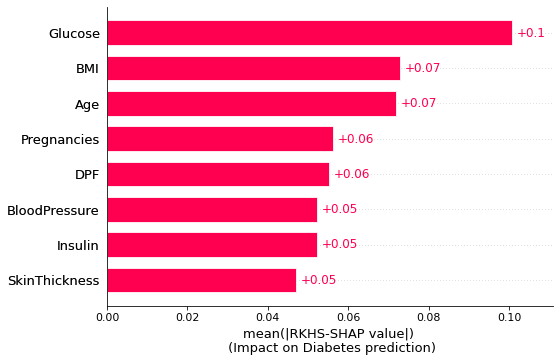}
    \caption{Beeswarm and barplot of the RKHS-SHAP values on the Diabetes for pima indian heritage dataset}
    \label{fig: diabetes_india}
\end{figure}

Figure~\ref{fig: diabetes_india} demonstrated how RKHS-SHAP explains the kernel logistic regression. The top predictor, "Glucose", which measures the plasma glucose concentration 2 hours in an oral glucose tolerance test, aligns with the intuition that it should be strongly predictive to whether a person is diabetic. Also, high BMI leading to someone more likely to be diabetic is also reflected from RKHS-SHAP values.

\begin{figure}[h!]
    \centering
    \includegraphics[width=0.55\textwidth]{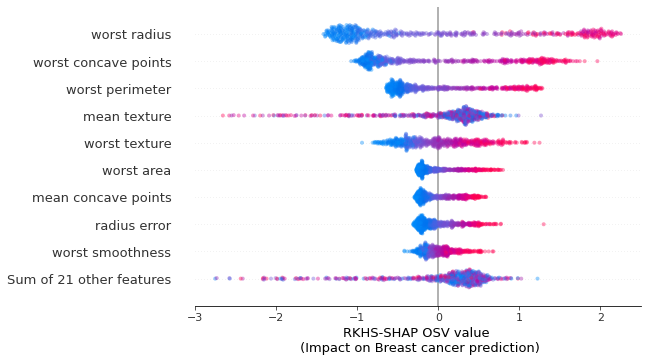}
    \includegraphics[width=0.44\textwidth]{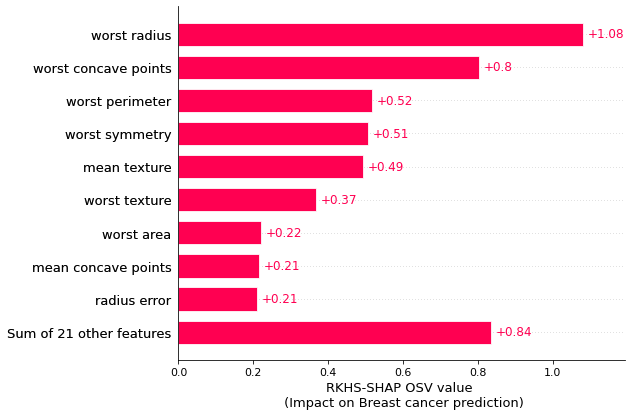}
    \caption{Beeswarm and barplot for the breast cancer prediction problem}
    \label{fig: cancer}
\end{figure}

\paragraph{Breast Cancer Classification} Next, we apply \textsc{RKHS-SHAP} to the breast cancer wisconsin dataset\footnote{https://goo.gl/U2Uwz2} to interpret the kernel logistic regression we have fitted to predict whether a patient might have breast cancer given their attributes. Features are computed from a digitized image of a fine needle aspirate (FNA) of a breast mass. They describe characteristics of the cell nuclei present in the medical image. There are $569$ data and $30$ features. When running \textsc{RKHS-SHAP}, we did not use all $2^{30}$ coalitions but subsampled $10000$ coalitions instead. Convergence analysis of such an approach is studied extensively by \cite{covert2021improving}, where they empirically show that the algorithm will converge in $\cO(n)$. Results are shown in Figure \ref{fig: cancer}. We can see that features such as "worst radius", "worst concave points", "worst perimeter" that describes the cell nuclei present in the breast mass, are most predictive to whether a patient has cancer or not. 

\paragraph{Census Income dataset} In the following, we will explain the kernel logistic regression deployed to predict the probability of an individual making over \$ 50K a year in annual income using the standard UCI adult income dataset. There are 48,842 number of instances and 14 attributes. 
\begin{figure}[h!]
    \centering
    \includegraphics[width=0.54\textwidth]{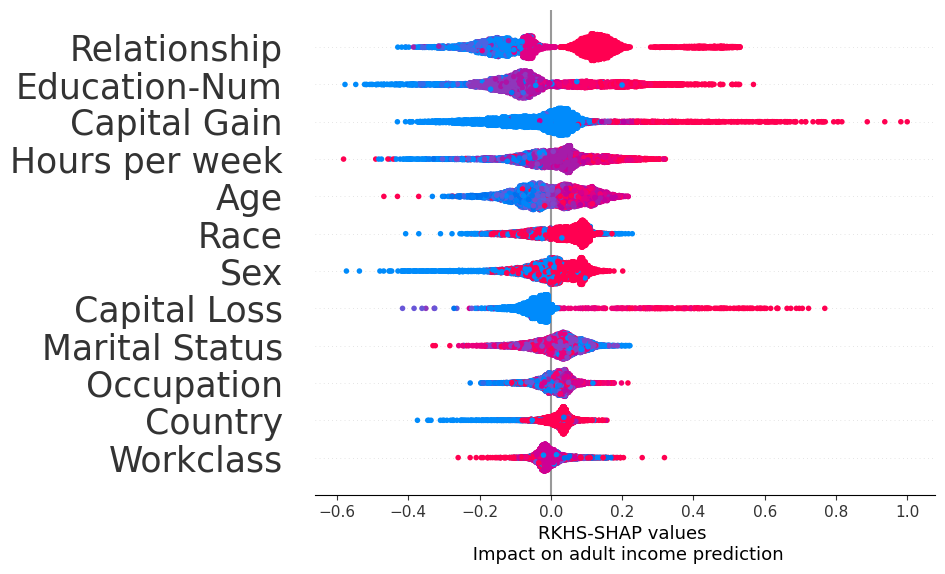}
    \includegraphics[width=0.45\textwidth]{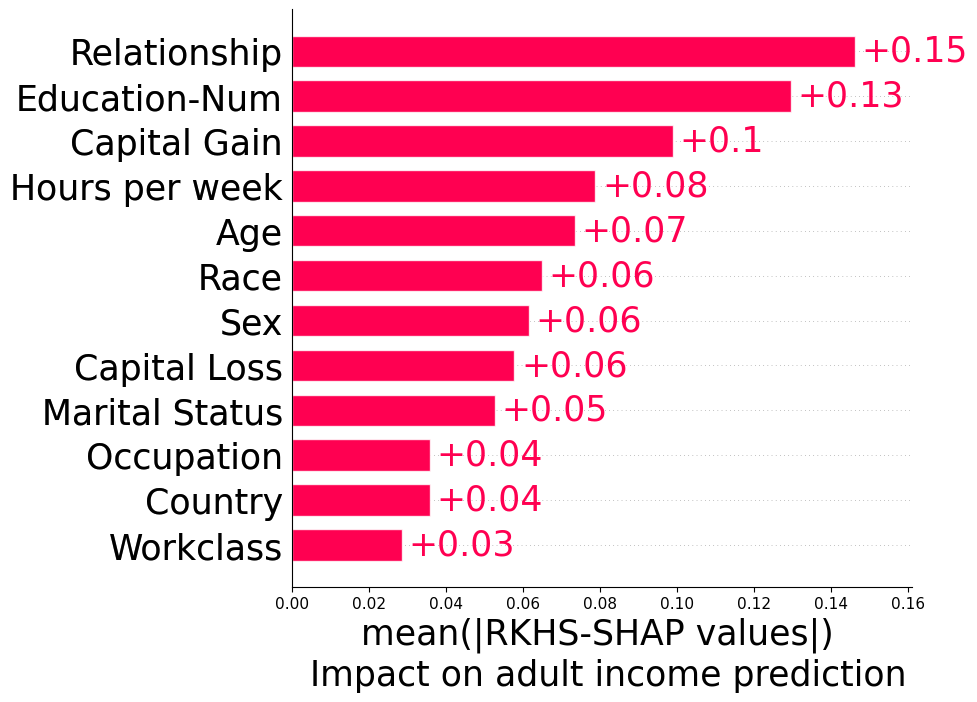}
    \caption{Beeswarm and barplot for the census income data prediction problem}
    \label{fig:my_label}
\end{figure}
We see that features such as relationship, education level and capital gain are most predictive of whether a person earns more \$ 50k a year. We see that as a person grows older, it is more likely to earn more, but the effect is not as impactful as, e.g. Education level or Capital gain.

\begin{figure}[t!]
    \centering
    \includegraphics[width=\textwidth]{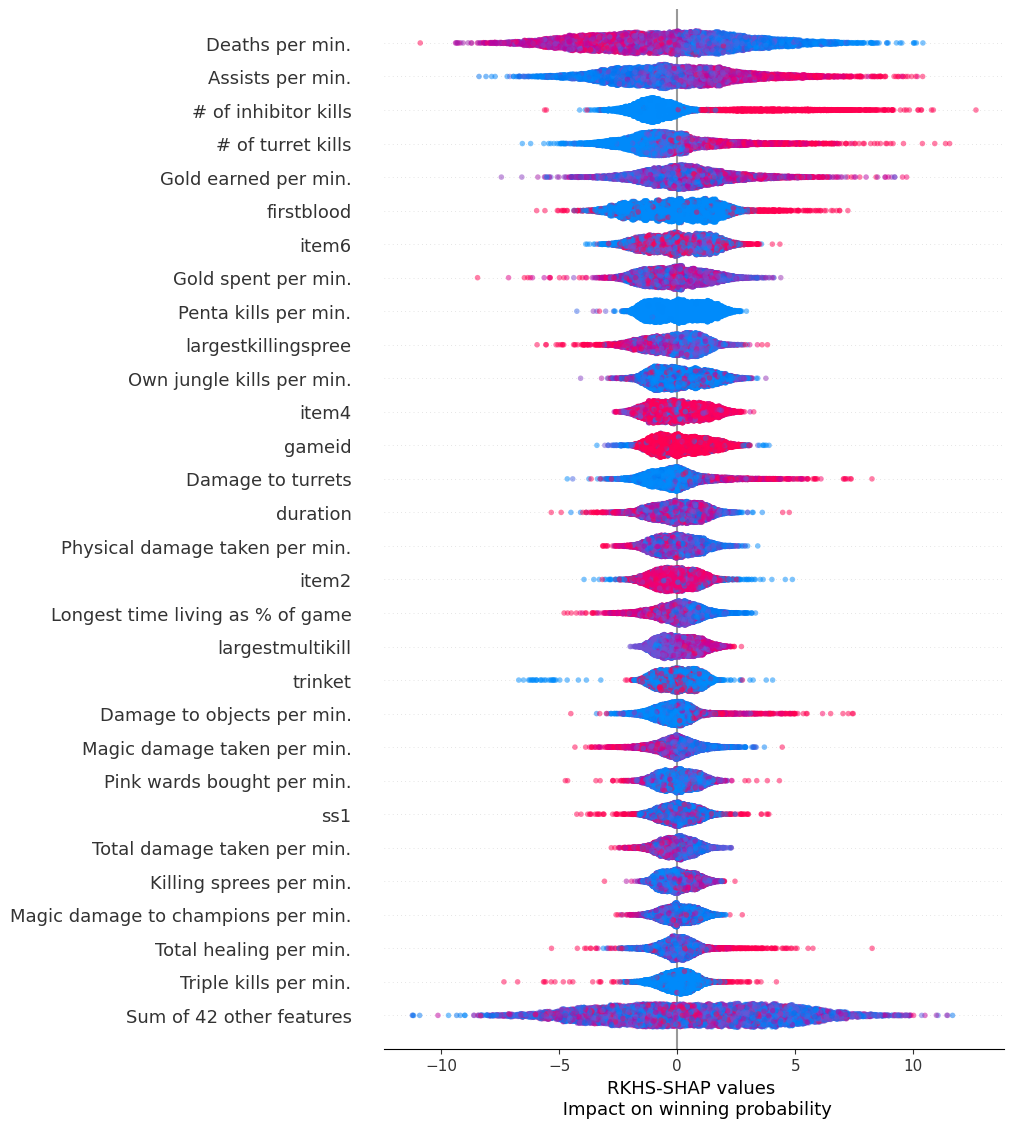}
    \caption{Beeswarm plot for the League of Legends player winning prediction problem obtained using RKHS-SHAP.}
    \label{fig:lol}
\end{figure}

\begin{figure}[t!]
    \centering
    \includegraphics[width=\textwidth]{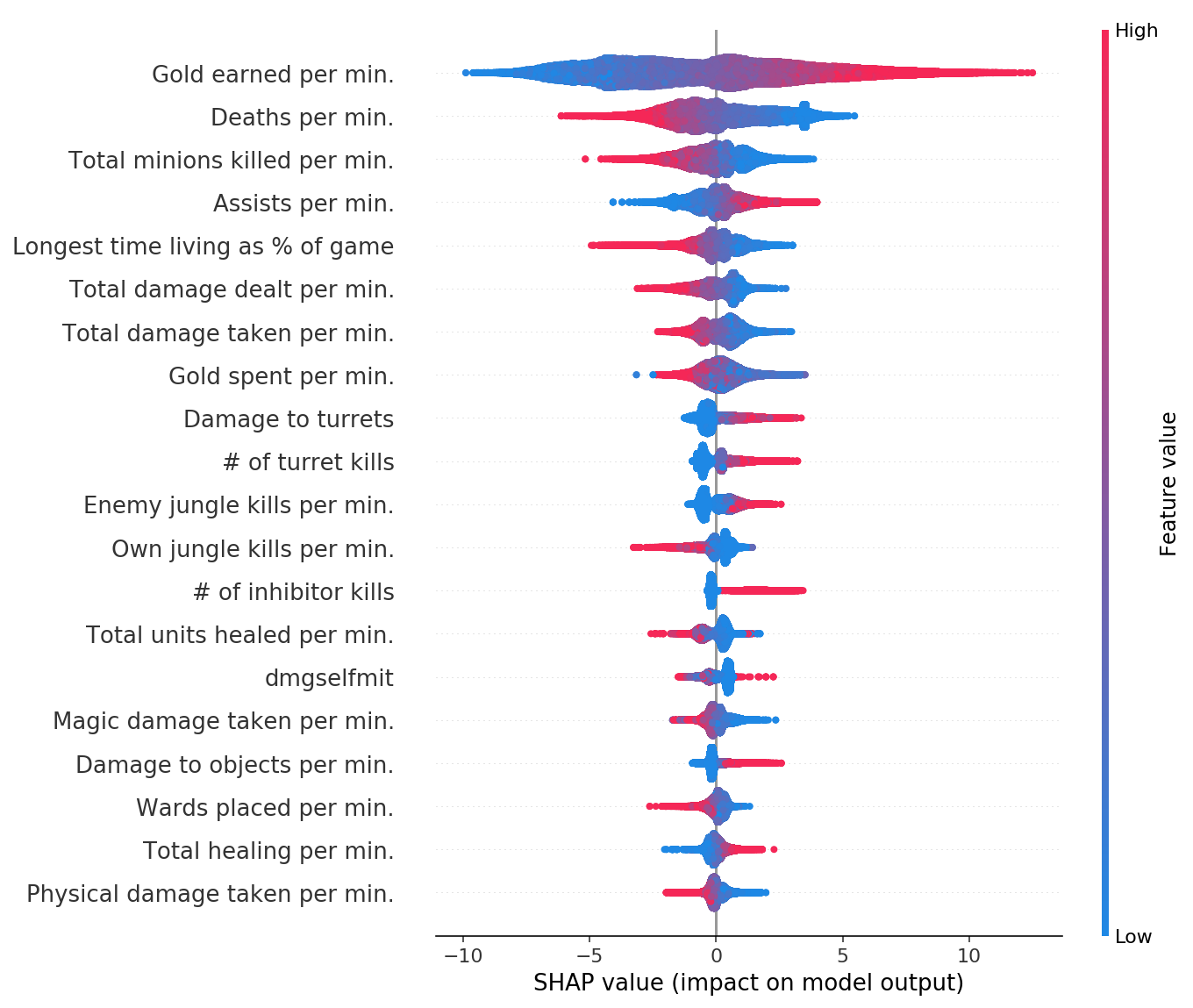}
    \caption{Beeswarm plot for the League of Legends player winning prediction problem obtained using TreeSHAP. Similar insights are recovered compared to RKHS-SHAP. However, since the two methods are explaining different models -- an RKHS function and a tree, it is not possible to tell which one gives more "correct" explanation.}
    \label{fig:lol_tree}
\end{figure}

\paragraph{League of Legends Win Prediction} Finally, we use the Kaggle dataset League of Legends Ranked Mathches which contains 1,800,000 players matches starting from 2014. We follow the preprocessing steps from~\cite{LOL_example}, and obtained $71$ features at the end. We deploy RKHS-SHAP to explain the fitted Kernel Logistic regression model and obtain results in Figure~\ref{fig:lol}. We see that features such as "Deaths per min" and "Assists per min" are most influential to the match outcome. It follows the game mechanism, as a player is intuitively considered as "strong" if he doesn't die often in a round of the game. We would also like to point out we recover similar explanations from~\cite{LOL_example}, where they applied TreeSHAP to recover the explanations, see Fig.~\ref{fig:lol_tree}. Interestingly, our kernel logistic regression seems to believe that "Gold earned per min" is less informative to the winning probability compared to "Deaths per min", which is different to the results obtained from the tree ensembles. 

